# Towards Sustainable Learning in Online Education: A Reinforcement Learning Approach

Chaofan Zhai

*Carlson School of Management, University of Minnesota*

Yicheng Song

*Carlson School of Management, University of Minnesota*

Ravi Bapna

*Carlson School of Management, University of Minnesota*

Junyao Ye

*MaiMemo Inc.*

**Abstract**

Online education offers unprecedented scalability and accessibility to global learners from diverse backgrounds, but it often suffers from low engagement and poor long-term learning effectiveness. To address these challenges, we introduce AI-Tutor, areinforcement learning–based model designed to promote sustainable learning by optimizing both short- and long-term learning outcomes. In the short term, AI-Tutor draws on cognitive theory to guide learners through a balance of acquiring new knowledge and reinforcing prior learning. In the long term, it models learner engagement to inform strategies that sustain motivation and reduce dropout. These enhancements enable AI-Tutor to provide personalized guidance that fosters both effective learning and sustained participation. Empirical evaluations on 23 million learning records from 33,700 learners show that AI-Tutor consistently outperforms state-of-the-art baselines across engagement, knowledge retention, and final learning outcomes. Learning path analyses further reveal how AI-Tutor adapts its strategies to learners with diverse profiles, offering adaptive and human-centered support.

1

## 1. Introduction

Online education is rapidly transforming global learning, reaching hundreds of millions of learners. Its market is projected to grow from $34 billion in 2024 to $682 billion by 2033 (IMARC 2024). Offering flexible and affordable access to high-quality content, it enables diverse learners to pursue academic, professional, and lifelong learning goals. Despite this remarkable expansion, online education continues to face persistent challenges in sustaining learner engagement and ensuring long-term learning effectiveness (Lockee 2021). The completion rate for MOOCs remains low and has declined over time, dropping from approximately 6% in 2014–15 to 3.13% in 2017–18 (Reich and Ruiperez-Valiente 2019). Meanwhile, learners in online settings often earn lower grades than´ their peers in in-person classes (Jack et al. 2023), and long-term knowledge retention suffers due to the lack of interactive engagement and structured review (Altindag et al. 2024). These findings highlight a fundamental tension in online education: while it expands access at an unprecedented scale, it often struggles to deliver sustained engagement and durable knowledge retention.

To ensure the healthy and sustained growth of online education, increasing attention is being directed not only toward broadening access but also toward fostering deeper and more durable learning outcomes. In this context, *sustainable learning* (Peris-Ortiz and Lindahl 2015) offers a valuable framework for evaluating long-term educational success. It emphasizes the design of engaging and effective learning experiences that yield lasting cognitive and motivational benefits for learners. Two key dimensions define this framework: sustained engagement, which reflects learners' ongoing motivation and active participation, and long-term effectiveness, which ensures that acquired knowledge is retained over time. A promising strategy for supporting sustainable learning is the use of personalized learning systems. These systems go beyond static content delivery by dynamically adjusting recommendations and feedback based on each learner's evolving knowledge state and engagement level. This adaptability enables more motivating and contextsensitive learning experiences that help keep students on track and foster deeper understanding. Prior research has demonstrated that personalized tutoring systems can significantly improve both learning performance and retention compared to one-size-fits-all instruction (Kumar and Mehra 2024). As such, aligning instructional content with individualized learning trajectories represents a critical direction for advancing sustainable learning in online education.

To support this goal, AI-powered personalized learning systems have been increasingly adopted in online education to enhance the effectiveness of instructional delivery and learner support (Li et al. 2024, Wang et al. 2025). Existing research in this area generally falls into two streams. The first stream focuses on predictive models that aim to identify the next learning module a student is most likely to complete successfully. These models often leverage graph neural networks (GNNs) to capture structural relationships among knowledge concepts—such as prerequisites or semantic dependencies—thereby improving the modeling of student learning trajectories (Tong et al. 2020). Other approaches in this stream apply sequential modeling techniques to capture the temporal evolution of a student's knowledge state (Xiong et al. 2016). While effective, these methods tend to focus on short-term objectives, such as maximizing the probability of success on the next learning module. In contrast, the second stream adopts a Reinforcement Learning (RL) perspective (Wang et al. 2020), shifting the emphasis from short-term gains to long-term learning rewards. Rather than focusing on immediate success probabilities, RL-based methods optimize the entire learning path to maximize cumulative learning outcomes over time (Cao and Leng 2021). The RL paradigm
is particularly well-suited to capturing the sequential, goal-directed nature of learning and aligns naturally with the broader objective of promoting sustainable learning in online education.

While recent RL approaches have made promising strides in optimizing long-term learning outcomes, two critical challenges remain largely unaddressed. First, most RL-based tutoring systems implicitly assume that students will consistently follow personalized recommendations and persist through the entire course. In practice, however, early disengagement and dropout are widespread in online learning environments (Xu and Jaggars 2014), substantially undermining the long-term benefits these systems are designed to deliver. Thus, it is critical to manage the trade-off between maintaining learner engagement and promoting academic progress. On one hand, overly aggressive strategies that continuously push learners forward without accounting for their cognitive load, motivation, or psychological readiness can lead to frustration, anxiety, and eventual withdrawal (Tyler-Smith 2006). On the other hand, excessively simplistic or repetitive content may bore learners and reduce their motivation to continue (Rezaee and Seyri 2022). Therefore, fostering sustainable learning in online environments requires carefully balanced recommendation policies—ones that adaptively calibrate content not only to drive academic progress but also to support continued engagement. Second, most RL-based systems focus primarily on accelerating academic progression by recommending new learning modules. Yet learning is not a one-way trajectory—successfully mastering a module once does not ensure long-term retention. There exists an inherent trade-off between knowledge expansion and retention. Without systematic knowledge review and reinforcement, learners are likely to forget earlier material, a phenomenon well documented in cognitive psychology (Ebbinghaus 2013). Thus, tutoring strategies that emphasize constant forward momentum without mechanisms for review may fall short of fostering durable knowledge retention.

To address these gaps, we develop an RL framework that jointly considers learner engagement and long-term knowledge retention. We propose a personalized AI-Tutor built upon an RL framework designed to align with the dual goals of sustainable learning. The first goal of our approach is to balance learner engagement and academic progress. Rather than treating students as passive recipients of instruction, we model them as active participants who voluntarily choose whether to persist based on their learning experience. To support this behavior, our RL framework explicitly incorporates engagement probability into the long-term reward estimation. This encourages the system to recommend learning paths that foster positive and motivating experiences. As a result, AI-Tutor dynamically adapts to each learner's profile, maintaining an optimal level of challenge that cultivates both a sense of progress and intrinsic satisfaction. The second goal is to manage the trade-off between knowledge expansion and retention. The model must balance recommending new content—which extends the learner's knowledge frontier—with review content, which strengthens long-term memory and deepens understanding. Our recommendation strategy is grounded in wellestablished learning and cognitive theories. For acquiring new concepts, we draw on constructivist learning theory (Von Glasersfeld 2012), which posits that learning is most effective when instructional content lies just beyond the learner's current knowledge boundary but remains attainable with appropriate guidance. For review scheduling, we build on the theory of forgetting curve (Finkenbinder 1913), which characterizes the temporal decay of memory and underscores the importance of spaced repetition for long-term retention. To achieve an appropriate balance between knowledge expansion and knowledge retention, we model the gains from both side and jointly integrate them into the reward function of the RL framework. With these two advancements to the RL framework, AI-Tutor aims to learn tutoring strategies that are well aligned with the core objectives of sustainable learning. Furthermore, to ensure that the learner state is accurately represented and that recommended actions are both structurally coherent and pedagogically effective, we incorporate a knowledge graph that models the relationships among knowledge items. This module enables the AI-Tutor to precisely track learners' evolving states and to construct a context-aware action space for recommending. To empirically evaluate AI-Tutor, we apply it to the domain of online language learning. Experimental results on 23 million learning records from 33,700 learners demonstrate that the proposed model outperforms state-of-the-art baselines in course completion rates, final test scores, and knowledge coverage rates. These empirical findings underscore the strength of AITutor's personalized learning guidance in achieving both effective long-term knowledge retention and sustainable engagement, advancing the dual goals of sustainable learning in online education.

We review the related literature in Section 2, identifying key research gaps and positioning the contributions of this study. Section 3 introduces the AI-Tutor, a reinforcement learning model grounded in cognitive theory and designed to personalize learning paths that optimize long-term learning outcomes. In Section 4, we evaluate the AI-Tutor using a large-scale dataset from a leading language-learning platform, demonstrating its superiority over state-of-the-art baselines in fostering sustainable learning. Section 5 further investigates the learning trajectories generated by the AI-Tutor, comparing them with those produced by alternative personalized learning systems. This analysis reveals the core strategies the model employs to promote sustainable learning and illustrates how it dynamically adapts personalized recommendations based on individual learner profiles and behavioral patterns. Finally, Section 6 concludes the paper and outlines several avenues for future research and development.

## 2. Related Work

Our research is closely related to three streams of literature. The first focuses on predictive models designed to enable personalized learning by estimating student performance on the next recommended knowledge item. The second stream encompasses RL approaches for online education and personalized learning path planning, which aim to optimize instructional sequencing based on long-term learning outcomes. The third stream examines spaced repetition methods, which reinforce knowledge retention by scheduling review at strategically timed intervals.

### 2.1. Predictive Models for Personalized Learning

The first stream of related literature focuses on enhancing predictive modeling for personalized learning through knowledge-aware architectures. A central insight in this work is that learning content is inherently structured—knowledge concepts are linked through prerequisite, semantic, or cognitive relationships. To leverage these dependencies, researchers have increasingly integrated structural information into predictive models to improve student performance prediction.

Graph neural networks (GNNs) have been widely adopted to capture the relational structure among knowledge concepts. Nakagawa et al. (2019) introduce a graph-based knowledge tracing model that incorporates prerequisite relationships to enhance prediction accuracy. Yang et al. (2020) further propose Graph-based Interactive Knowledge Tracing, combining GNNs with memory modules to model long-term student behavior. Liu et al. (2020) extend this line by jointly modeling concept structures and student interaction histories, highlighting the value of structural and sequential integration. In parallel, sequence-based models emphasize the evolving nature of student knowledge. Piech et al. (2015) pioneer this direction with Deep Knowledge Tracing (DKT), using recurrent neural networks to model learning trajectories. Pandey and Karypis (2019) build on this by introducing a self-attentive model that better captures dependencies in learning sequences.

These predictive modeling approaches can facilitate personalized content recommendations and performance forecasting. However, they do not fully address key limitations related to longterm learning outcomes—such as sustaining learner engagement over time, promoting knowledge retention, and balancing the trade-off between knowledge expansion and retention.

### 2.2. Reinforcement Learning Models for Personalized Learning

The second stream of related literature focuses on applying RL to personalized education. Unlike predictivemodelsthatpassivelyestimateoutcomesbasedonhistoricaldata,RLframeworksactively learn optimal teaching strategies through interaction, using learner feedback in the form of rewards. For example, the system may receive a positive reward when a student answers a question correctly and a smaller or no reward when the response is incorrect. This paradigm allows the system to make sequential, goal-directed decisions that adapt over time to maximize long-term learning outcomes, making RL particularly well-suited to adaptive learning environments.

Early work by Raghuveer et al. (2014) demonstrated the feasibility of using reinforcement learning (RL) to personalize content sequencing in intelligent tutoring systems. Building on this foundation, Cai et al. (2019) introduced a contextual bandit model that tailors question selection to a student's ability, offering a scalable solution with strong empirical performance. Expanding beyond discrete selection, Wang et al. (2020) developed a policy optimization method using entropy regularization to balance exploration and exploitation in curriculum design. In another extension, Bassen et al. (2020) introduced an RL agent designed to select the most suitable instructional format for each educational activity, aiming to optimize learning outcomes while minimizing the number of knowledge items assigned. Most recently, Xiao and Wang (2024) proposed an RL framework designed to determine the optimal learning intervals.

Despite these advances, existing RL-based approaches often face two key limitations in the context of sustainable learning. First, most reward functions prioritize knowledge expansion while overlooking the benefits of review and knowledge consolidation. Second, these models typically assume stable learner engagement and do not explicitly model dropout risk in value estimation. As a result, they may overestimate the expected return of certain learning paths and fail to strike an effective balance between promoting new knowledge and reinforcing prior learning.

**2.3. Spaced Repetition Models for Knowledge Retention**

The third stream of related literature focuses on enhancing knowledge retention through cognitive theory–driven approaches. A central focus in this area is on spaced repetition models, which aim to optimize the timing of content reviews based on psychological insights into memory retention. Grounded in Ebbinghaus's forgetting curve (Finkenbinder 1913, Radvansky et al. 2022), this approach posits that memory decays exponentially over time but can be strengthened through strategicallytimedreviewsessions.Byspacinglearninginteractionsappropriately,spacedrepetition helps consolidate knowledge into long-term memory and mitigate the effects of forgetting.

Early foundational work by Dempster (1989) highlighted the robust empirical effects of spacing on memory performance, advocating its underutilized potential in educational settings. Building on this, Pavlik Jr and Anderson (2005) developed a computational model that quantifies the influence of prior exposure and spacing on recall probability, providing a principled way to schedule practice for improved retention. Sense et al. (2016) further introduced individualized retention models, emphasizing that optimal review intervals vary across learners. Settles and Meeder (2016) proposed aspacedrepetitionsystembasedonstatisticalmodelstooptimizeschedulingpolicies,demonstrating enhanced flexibility and personalization. More recently, Ye et al. (2022) advanced a stochastic modeling approach that accounts for uncertainty in memory retention and enables adaptive review recommendations. Extending these ideas, Su et al. (2023) formulated the scheduling problem as a stochastic shortest-path task, providing a unified framework that integrates memory prediction with optimal review planning.

Despite their effectiveness in modeling individual memory dynamics, existing spaced repetition approaches face notable limitations. First, these models typically schedule item reviews based solely on each item's past performance, treating items independently. This item-level focus neglects structural relationships among knowledge items, including pedagogical dependencies that shape effective learning sequences and influence long-term retention. Second, most spaced repetition models adopt a reactive strategy: they trigger reviews only after a learner's memory strength for a given item has already declined. This reactive approach lacks a proactive mechanism to manage memory strength holistically across the entire set of knowledge items. As a result, these methods may fall short in systematically achieving long-term knowledge retention.

Insummary,weproposeAI-Tutorforsustainableonlineeducationthatuniquelyintegratesknowledge structure, learner engagement, and memory retention. This work makes two key contributions:

**Figure 1** **Personalizing Learning Paths via the AI-Tutor**

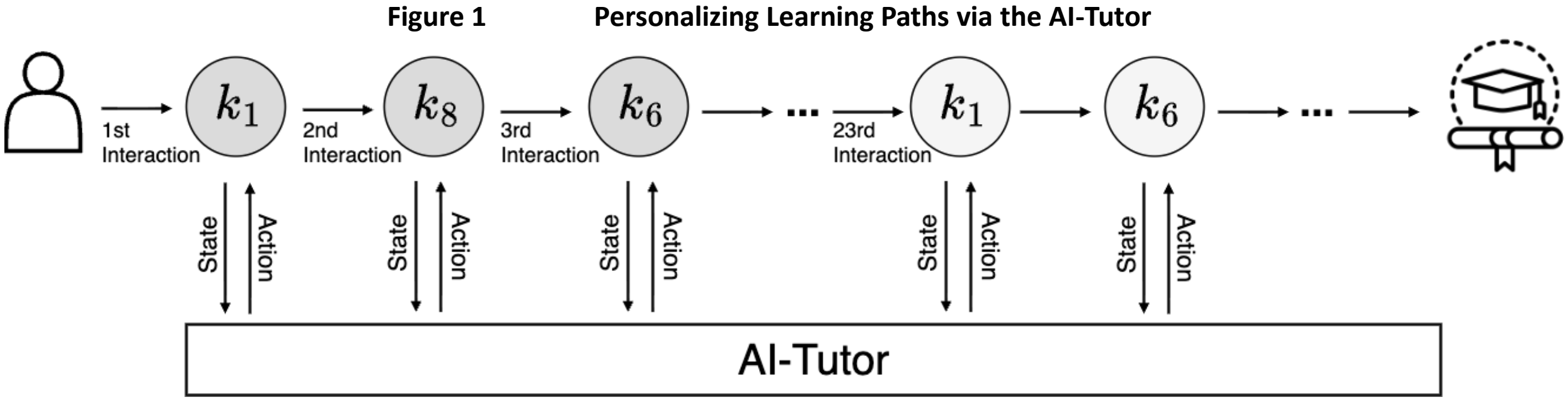

*Note.* An exemplary user interacts with the AI-Tutor. At each interaction, the AI-Tutor observes the learner's state—such as knowledge mastery and engagement level—and selects an action by recommending the next learning item. The shading of each circle reflects the learner's retention of that item; items with lower retention (lighter gray) may be reviewed to strengthen understanding.

1) **A Novel RL Framework for Sustainable Learning:** Our framework explicitly balances knowledge expansion and retention while incorporating behavioral engagement, addressing limitations in prior work that overlook these dimensions. 2) **Theory-Driven Model Design:** Grounded in cognitive and memory theories, our approach is among the first to embed domain theories directly into the RL model formulation for online education applications. This enables a theory-aligned and effective solution for a personalized learning tutor. In the following section, we introduce the design and implementation of AI-Tutor, detailing how these components are operationalized.

## 3. AI-Tutor

Let's assume a student takes an online course with a knowledge pool $K = \{k_1, \ldots, k_n\}$, where $k_i$ denotes the $i$-th knowledge item. At each time step $t$, the student engages with a knowledge item in $K$, which may involve learning a new concept or reviewing previously studied material. The ultimate goal is for the student to complete the course by fully mastering all knowledge items in $K$. Importantly, students may choose to drop out at any point, resulting in early termination.

As illustrated in Figure 1, the AI-Tutor is a personalized learning-path recommender system. It begins by recommending the knowledge item $k_1$ to the student, denoted as the first action $a_1$. After receiving feedback, it proceeds to recommend the next item, for example $a_2 = k_8$, and continues this process until the student either completes the course or disengages. Mastery of $k_1$ in the first interaction does not guarantee long-term retention. Therefore, the AI-Tutor prompts a review of $k_1$ again in the 23rd interaction. The goal of the AI-Tutor is to guide the learner through the knowledge pool in a way that supports course completion and enhances long-term retention, thereby promoting sustainable learning. This setting is naturally formulated as an RL problem, where the AI-Tutor acts as an RL agent that sequentially interacts with the learner by recommending knowledge items to optimize long-term learning outcomes. However, applying standard RL methods to this context introduces several key challenges:

1. Unlike typical RL settings where actions are independent, knowledge items in the knowledge pool $K$ are inherently interconnected. These structural dependencies significantly affect learning outcomes and must be considered when recommending learning paths. Furthermore, the learner's state is not adequately captured by a static representation; instead, it is best modeled as a sequence of previously encountered knowledge items that reflect the learner's evolving understanding. Thus, it is essential to structurally represent knowledge items for action recommendation and state learning.

2. Standard RL approaches typically assume that the environment responds in a stable and predictable manner to the agent's actions. In the context of education, this would imply that studentsremainconsistentlyengagedandcontinueinteractingwiththesystemasintended.However, learner behavior is inherently stochastic and influenced by numerous unobserved factors, such as motivation, cognitive fatigue, or external distractions. As a result, the assumption of a fully cooperative and persistent student does not hold, and many learners disengage before completing the course. This highlights the importance of managing the trade-off between learning progress and sustained engagement, which is often overlooked in existing RL studies in educational settings.

3. Existing RL-based educational systems typically prioritize course progression by recommending new knowledge items. Yet learning is not a unidirectional process. Learning a knowledge item once does not guarantee its retention over time. Without a structured review, students are likely to forget previously learned knowledge. This introduces an additional trade-off between *knowledge expansion* (acquisition of new material) and *knowledge retention* (consolidation of prior learning).

To address these challenges, we propose several key enhancements to the conventional RL framework. To address the first challenge, we model knowledge items within a knowledge graph as described in Section 3.1. This graph serves as a rich relational representation of the learning space and is used to construct both the learner's state and the set of pedagogically appropriate actions. Second, informed by memorization theory in cognitive psychology (Finkenbinder 1913, Radvansky
etal.2022),wedecomposetheimmediaterewardoflearningaknowledgeitemintotwocomponents: the learning gain from knowledge acquisition and knowledge retention, as described in Section 3.2.
Thisrewardformulationencouragesthemodeltostrikeabalancebetweenacquiringnewknowledge and reinforcing learned knowledge. Third, we explicitly model the learner's engagement probability over time and revise the classical Bellman equation to incorporate the diminishing impact of future rewards under the risk of early dropout, as detailed in Section 3.3. These advancements guide the AI-Tutor to recommend content that promotes both long-term retention and sustained participation. Next, we introduce each component of the AI-Tutor in detail. A complete list of notations used throughout the paper is provided in Appendix A.

### 3.1. Knowledge Graph

A Knowledge Graph (KG) captures the structure of a knowledge domain by representing individual knowledge items as nodes and their relationships, such as prerequisites, semantic similarity, or conceptual dependencies, as edges (Wang et al. 2017). Such a graph representation allows the system to model not only isolated knowledge items but also the logical and pedagogical connections between them. By explicitly encoding these relationships, the KG provides a structured foundation that supports AI-Tutor to learn optimal representations of knowledge items and learner states.

Inthisstudy,weaimtoconstructaknowledgegraphforanonlinecoursetoenhancetheAI-Tutor's ability to recommend effective and personalized learning sequences. The nodes correspond to all knowledge items in the set $K = \{k_1,..., k_n\}$. Edges are defined based on the principle that mastering one knowledge item supports or facilitates the learning of another. Intuitively, such relationships can be derived from two complementary sources. When there are clear structural dependencies among knowledge items (e.g., teaching algebra before calculus), the relationships can be directly extracted from curricula. In cases where such dependencies are not explicitly defined, we infer them empirically from learners' historical learning experiences to uncover latent support relationships between knowledge items. To construct edges that are both pedagogically grounded and empirically informed, we incorporate the following two complementary types of edges into the KG:

***Curriculum-Driven Edge:*** $e_{i,j}^{cd}$ captures prerequisite or instructional dependencies grounded in the curriculum structure of the online course (Bauman and Tuzhilin 2018). These edges represent the intended conceptual progression of learning—what should be mastered as prerequisites before advancing to subsequent concepts. For instance, in language learning, understanding the simple present tense may be a prerequisite for mastering the present perfect tense. Such edges are typically constructed by domain experts or derived from the curriculum structure. They ensure that the AI-Tutor respects the interdependent nature of knowledge acquisition and avoids recommending content that is too advanced without foundational understanding.

***Empirically-Driven Edge:*** $e_{i,j}^{ed}$ is extracted from historical learning experience. Inspired by association rule mining (Ghoshal and Sarkar 2014), a technique for uncovering latent relationships between items in large datasets, we aim to identify pairs of knowledge items—denoted as $i$ and $j$—such that mastery of item $i$ increases the likelihood of mastering item $j$. We define *support* as the proportion of learning interactions in which both item $i$ and item $j$ were mastered. *Confidence* refers to the proportion of interactions where item $j$ was mastered among those in which item $i$ had already been mastered; in other words, it reflects how frequently mastery of item $j$ follows mastery of item $i$ during learning interactions. To assess whether this relationship exceeds random chance, we calculate *lift*, defined as the ratio of *confidence* to the marginal probability of mastering item $j$ based on the *support*. A lift greater than 1 indicates a positive dependency, which suggests that mastering item $i$ facilitates mastery of item $j$. When such a dependency pattern is identified (Aggarwal et al. 2009), we add a directed edge $e_{i,j}^{ed}$ from knowledge item $i$ to item $j$ in the KG.

**Figure 2** **A Knowledge Graph Example**

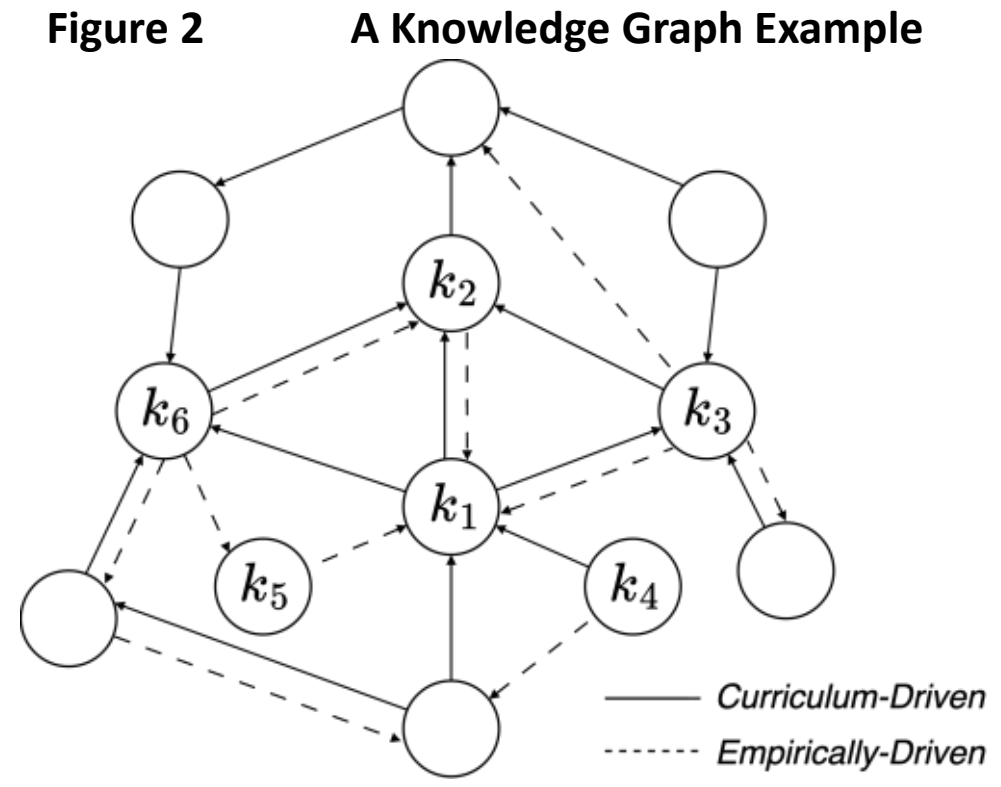

By integrating curriculum- and empirically-driven edges into a unified graph structure, the AI-Tutor leverages both pedagogically grounded and empirically informed effectiveness to guide personalized learning paths. Accordingly, we define the knowledge graph as a directed graph $G = (V, E)$, as shown in Figure 2. Each node $k_i \in V$ represents a knowledge item from the knowledge pool $K$. The edge set $E \subseteq V \times V$ captures dependencies among knowledge items and consists of two types of directed edges: curriculum-driven edges $e_{i,j}^{cd}$ and empirically-driven edges $e_{i,j}^{ed}$.

Given graph $G$, we employ a Graph Neural Network (GNN) to learn embeddings of both knowledge items and the learner's knowledge state. GNNs are designed to operate on graph data by iteratively updating each node's representation based on its own features and the aggregated information from its neighbors (Lyu et al. 2021, Ma et al. 2024). In our context, a key advantage of using a GNN is its ability to model both local and global patterns within the KG. This includes intrinsic properties of individual knowledge items and structural relationships such as prerequisites or learning hierarchies. For each item $k_i \in V$, the GNN aggregates information from the node itself and its neighbors to compute a $d$-dimensional embedding $\mathbf{u}_{k_i} \in \mathrm{R}^d$. This embedding can be used to represent a knowledge item $k_i$ that captures not only its intrinsic properties but also its contextual and structural roles within the graph. To represent a student's knowledge state before the $t$-th interaction, we extract the subgraph $g_t \subseteq G$, induced by the set of items previously learned by the student. We then obtain a $d$-dimensional subgraph embedding $\mathbf{u}_{g_t} \in \mathrm{R}^d$ by aggregating the embeddings of the nodes in this subgraph $g_t$. Specifically, in this study, we adopt Deep Graph Infomax (DGI) (Velickovic et al. 2019), an unsupervised GNN model, to learn embeddings of knowledge items and learner's knowledge states. DGI is particularly well-suited to our setting for several reasons. First, it is an unsupervised method that contrasts real subgraphs (e.g., real learning trajectories) with randomly corrupted ones to train the model to recognize coherent and pedagogically meaningful structures. This contrastive learning setup allows the model to retain informative structural patterns in the graph without requiring manual labels. Second, DGI excels at capturing both local and global graph structures, enabling it to embed each item with rich contextual awareness—such as its position in learning hierarchies and dependencies among concepts.

Beyond the above tasks, the KG also plays a critical role in constructing the action space for the RL model, ensuring that recommended items are theoretically grounded and empirically effective. This integration is further detailed in the following section.

### 3.2. AI-Tutor with RL

The RL formulation models the learning environment through three key components: *state*, *action*, and *reward*. Here, the *state* $s_t$ captures the learner's current condition prior to the $t$-th learning interaction; the *action* $a_t \in K$ corresponds to the knowledge item to be recommended at the $t$-th interaction; and the *reward* $r_t$ reflects the immediate learning gain derived from that interaction. Given these components, the RL agent aims to learn a value function $Q_\pi(s_t, a_t)$, which estimates the expected long-term reward when selecting action $a_t$ under the current state $s_t$, according to an action policy $\pi(s_t) = a_t$. This estimation considers not only the immediate reward $r_t$, but also the rewards that may be obtained in the future following the action policy $\pi$. Consequently, the agent is encouraged to learn an action policy $\pi$ that optimizes long-term reward $Q_\pi(s_t, a_t)$. Next, we formalize these key elements of the RL model in this setting.

***State*** is modeled as a dynamic representation of the learner's learning trajectory (Todri et al. 2020). Prior to the $t$-th interaction, the learner's knowledge state is captured by the embedding $\mathbf{u}_{gt}$ of the subgraph $g_t$ within the knowledge graph $G$, which is induced by the set of knowledge items the learner has previously learned. The target knowledge item in the $t$-th knowledge interaction $a_t$ is represented by the embedding $\mathbf{u}_{at}$ of the knowledge item $a_t \in G$. Additionally, we also incorporate a learning profile $\mathbf{lp}_{at}$ of the learner for the focal knowledge item $a_t$ up to time $t$, which includes: the cumulative number of times the learner has studied $a_t$, the time elapsed since the last interaction with $a_t$, whether the learner successfully recalled $a_t$ during the last interaction. Together, the learning interaction at time $t$ is represented as $\mathbf{l}_t = \{\mathbf{u}_{gt}, \mathbf{u}_{at}, \mathbf{lp}_{at}\}$, which encapsulates the learner's knowledge state, the target item embedding, and the learning profile. The learning trajectory up to time $t$ then can be represented as $L_t = \{\mathbf{l}_1, \ldots, \mathbf{l}_t\}$, which summarizes the learner's evolving knowledge trajectory. To encode this sequential trajectory into a compact and informative learner

state representation $s_t$, we adopt a Recurrent Neural Network (RNN) to process $L_t$:

$$s_t = \text{RNN}(L_t). \quad (1)$$

RNNs are well-suited to this setting due to their ability to model temporal dependencies and retain information across variable-length learning sequences. This allows AI-Tutor to track a learner's evolving knowledge state over time and adjust its recommendation policy based on the learning history. Unlike static representations, RNNs support a dynamic learner state that captures learning momentum, repetition patterns, and engagement fluctuations. The RNN component is trained endto-end as part of the AI-Tutor framework.

***Action*** $a_t \in K$ corresponds to selecting a knowledge item for the student to engage with at the $t$-th learning interaction. The objective of RL is to learn the action policy $\pi$ that chooses $a_t = \pi(s_t)$ for different learner state $s_t$ such that it maximizes the expected long-term reward. While the training of the actor policy $\pi(s_t)$—which selects $a_t$ given the learner's current state $s_t$ will be discussed in Section 3.3, our focus here is on designing the feasible action space. In principle, the action $a_t$ could be chosen from the entire knowledge pool $K$. However, such an unconstrained action space poses two key challenges. First, the large cardinality of $K$ leads to inefficient action exploration and increased computational costs. Second, knowledge items in $K$ are not independent. Recommending content that lies far beyond a learner's current knowledge level may lead to inefficiency in learning and disengagement. Constructivist learning theory (Von Glasersfeld 2012) suggests that learning is most effective when new material is just beyond the learner's knowledge boundary but still accessible with guidance. This principle aligns naturally with the KG in Section 3.1, which encodes how mastery of one concept facilitates the learning of another. Accordingly, we define the action spaceattime$t$ basedonthelearner'slearningpath,thesubgraph $g_t \subseteq G$.Thesetofallowableactions includes adjacent nodes to $g_t$, which represent new knowledge items that are directly connected to previously learned content and are, therefore, pedagogically appropriate. It also includes nodes within $g_t$, which represent previously learned items that can be reviewed to reinforce retention and prevent forgetting. Such action space masking optimizes exploration and ensures that all recommended knowledge items are developmentally appropriate.

***Reward*** $r_t$ represents the immediate learning gain the student derives from the $t$-th interaction. Accordingly, we aim to construct a reward function that aligns closely with the principles of sustainable learning to balance two key objectives: knowledge expansion and knowledge retention. Knowledge expansion occurs when a student acquires a new or previously forgotten concept. Notably, revisiting previously learned material can also contribute to knowledge expansion if the learner has forgotten the content—reacquisition in this context reflects a new learning gain. Meanwhile, even when the learner is able to recall the content successfully, the act of review also supports knowledge retention by strengthening understanding and long-term memory. Based on this distinction, we define the immediate reward from a learning interaction to include two components: learning gain from knowledge acquisition and learning gain from knowledge retention. Thus, the

immediate reward $r_t$ can be formulated as:

$$r_t = r_t^{\text{acq}} + \beta r_t^{\text{ret}}, \quad (2)$$

where $\beta$ is the hyperparameter tuned by grid-search to balance two learning gains.

The term $r_t^{\text{acq}}$ measures the gain from acquiring new knowledge during the $t$-th learning interaction. For the recommended knowledge item $a_t$, we first estimate the probability that the student has already mastered and could recall this item before the interaction. Let $p_t^{\text{rec}}$ denote this predicted probability. To quantify the learning gain from acquiring knowledge, we define it as:

$$r_t^{\text{acq}} = 1 - p_t^{\text{rec}}. \quad (3)$$

The value $1 - p_t^{\text{rec}}$ reflects the improvement in the student's knowledge acquisition. Specifically, it captures the gain from the predicted pre-interaction recall probability $p_t^{\text{rec}}$ to a state where the knowledge is freshly learned and assumed to be perfectly mastered after the interaction. Since the next knowledge item is selected based on the action mask on the knowledge graph, the AI-Tutor ensures that all recommendations are developmentally appropriate and achievable. Accordingly, we assume that the student could acquire the item upon engagement.

Beyond acquiring a new knowledge item, reviewing learned knowledge can also contribute to the immediate reward of knowledge acquisition. According to well-established theories of memorization (Finkenbinder 1913, Radvansky et al. 2022)—as illustrated in Figure 3—the probability of recalling a piece of learned knowledge decreases over time after it is learned. Immediately following

**Figure 3 Forgetting Curve**

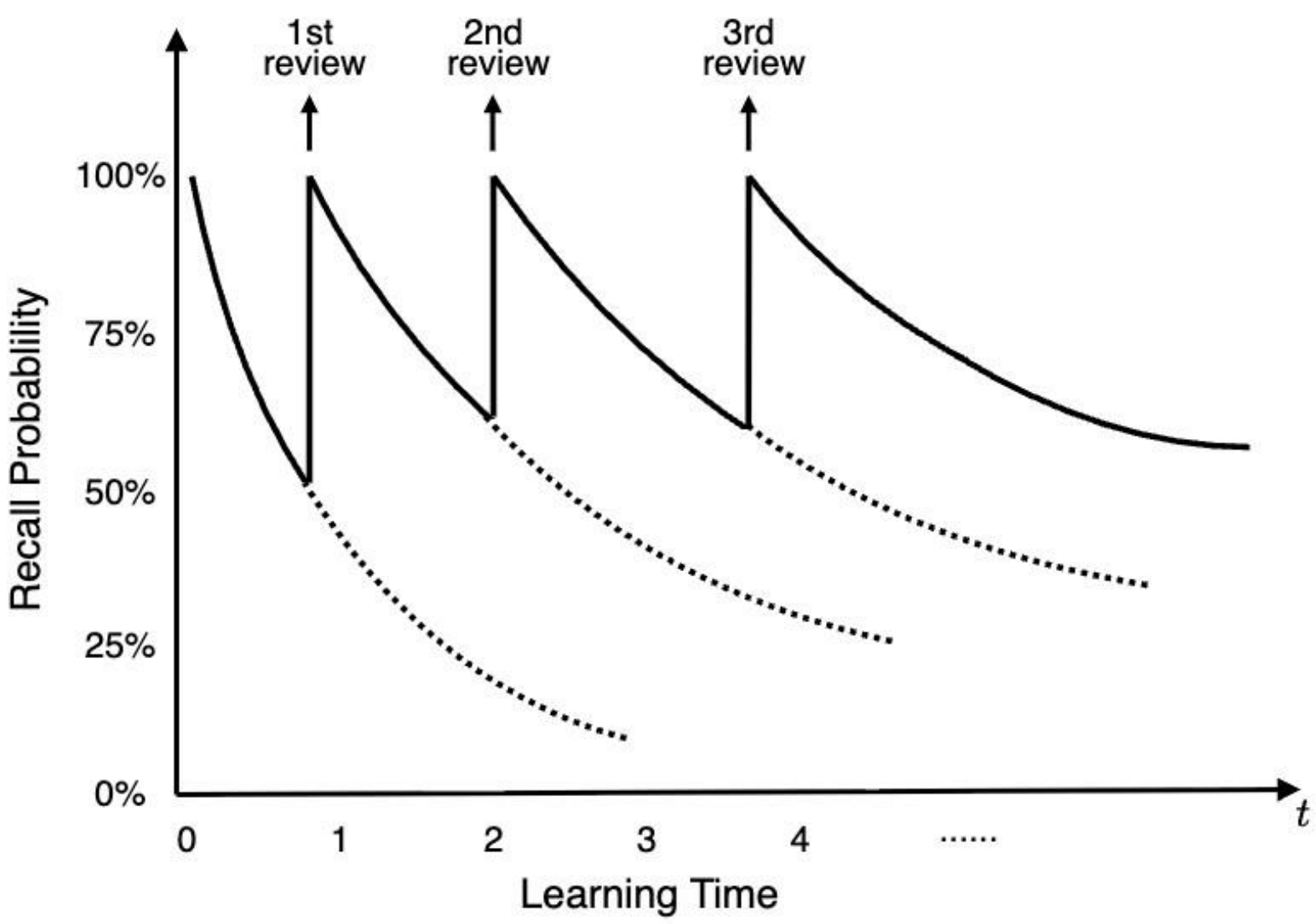

*Note.* The solid curve represents the forgetting (or recall probability) curve, which illustrates knowledge retention that decreases over time in our context. It is refreshed each time the knowledge is reviewed. The dashed curve represents how the curve would decrease if the knowledge were not reviewed. With an increase in reviews, the curve becomes progressively flatter.

the acquisition of a knowledge item $i$, the recall probability is assumed to be 100%. However, due to the effects of forgetting, the recall probability $p_t^{\text{rec}}$ decays over time $t$, even if the knowledge was once mastered. When the learner revisits and reviews the knowledge item $i$, the recall probability is refreshed back to 100%. In this sense, reviewing previously learned content can also be viewed as a form of learning gain.

Beyond knowledge acquisition gain, reviewing learned content also contributes to knowledge retention by strengthening memory traces. Thus, we introduce $r_t^{\text{ret}}$ that measures the gain in long-term knowledge strengthening after the $t$-th learning interaction. Theories of memorization (Finkenbinder 1913, Dempster 1989) have shown that repeated reviews flatten the forgetting curve of a knowledge item, thereby slowing the rate of forgetting over time. As illustrated in Figure 3, the decline in recall probability becomes significantly slower after multiple reviews, highlighting the cumulative effect of reinforcement on memory consolidation.

To quantify $r_t^{\text{ret}}$, we adopt the half-life decay model introduced by Settles and Meeder (2016), which models the recall probability as a function of the half-life memory strength and the elapsed time since the last review. Specifically, let $h_{pre(t)}$ denote the half-life strength of the knowledge item before the $t$-th interaction, and $\Delta t$ be the time interval since it was last learned/reviewed to the $t$-th interaction. The probability of recalling the knowledge item at the $t$-th interaction is then given by:

$$p_t^{\text{rec}} = 2^{-\Delta t / h_{pre(t)}}. \tag{4}$$

The half-life $h_{pre(t)}$ corresponds to the time it takes for the recall probability to fall to 50%. For example, when $\Delta t = h_{pre(t)}$, we have $p_t^{\text{rec}} = 0.5$. A larger $h_{pre(t)}$ indicates slower memory decay and thus stronger long-term retention. When the learner reviews the knowledge item in the $t$-th interaction, its half-life is updated from $h_{pre(t)}$ to $h_{post(t)}$, reflecting a strengthening of long-term memory. Therefore, we define the knowledge retention gain as the increase in half-life strength:

$$r_t^{\text{ret}} = h_{post(t)} - h_{pre(t)}. \tag{5}$$

With the explicit definitions of $r_t^{\text{acq}}$ and $r_t^{\text{ret}}$, the reward function in EQ 2 is now fully specified. Thisrewardformulationensuresthatthelearningpolicyisoptimizednotonlytofacilitateknowledge expansionbutalsotostrengthenlong-termknowledgeretention.Withtheaboveformulationof*State*, *Action*, and *Reward*, the RL framework is equipped to learn an optimal policy for a personalized tutor. In the following sections, we elaborate on the policy optimization in Section 3.3, and a detailed estimation of $p_t^{\text{rec}}$, $h_{pre(t)}$, and $h_{post(t)}$ in Section 3.4.

### 3.3. RL Policy Optimization

We train the AI-Tutor using the Actor-Critic (AC) framework. In the AC framework, the learning agent consists of two main modules: (1) a *critic network*, which estimates the expected long-term reward of taking a given action in a given state under a specific policy; and (2) an *actor network*, which learns a policy that selects actions under different states to optimize the long-term rewards. The actor iteratively refines its policy using feedback from the critic, while the critic is simultaneously updated to improve its long-term reward estimates based on new experiences collected from the actor's interactions with the environment. We adopt the AC framework because it combines the strengths of value-based and policy-based RL: it provides greater training stability, improved sample efficiency, and the flexibility to handle large and structured action spaces (Grondman et al. 2012), all of which are critical in the context of a personalized learning system.

To model long-term rewards, the classical Q-function in the critic network is defined recursively using the Bellman equation:

$$Q_\pi(s_t, a_t) = r_t + \gamma \mathbb{E}_{s_{t+1}, a_{t+1} \sim \pi} [Q_\pi(s_{t+1}, a_{t+1})] \,, \quad (6)$$

where $Q_\pi(s_{t+1}, a_{t+1})$ estimates the expected cumulative reward from future steps under policy $\pi$, and $\gamma \in [0,1]$ is a discount factor that balances present and future rewards. Typically, $\gamma$ is set to a fixed value (e.g., 0.99), and the RL agent aims to learn the optimal action $a_t$ for each state $s_t$. This formulation has been widely adopted in domains such as robotics (Kalashnikov et al. 2018), video games (Mnih et al. 2015), and finance (Ning et al. 2021), where maximizing long-term rewards is critical. The fixed discount factor $\gamma$ reflects the diminishing value of future rewards—due to environmental uncertainty in robotics and video games, or inflation in financial applications. Under such a mechanism, there is an implicit assumption that the environment is cooperative and that future discounted rewards are always attainable. To prevent the need to compute rewards over an infinite horizon, these environments typically define clear terminal states. These terminal states are often deterministic and directly triggered by the agent's actions under given states. For example, a robotic agent may end an episode by knocking over a cup; a game-playing agent may terminate the game by defeating all enemies or being defeated; and in finance, an investment horizon may be predefined as three months, after which all positions are closed. In such cases, the future reward term $Q_\pi(s_{t+1}, a_{t+1})$ is automatically set to zero by the environment's termination condition.

If we apply such a framework to online education, the classicalQ-functionimplicitlyassumes that students remain continuously engaged throughout the learning process, making future discounted rewards attainable until the course concludes. However, in practice, learners' engagement in online education is highly variable (Reich and Ruiperez-Valiente 2019); students may disengage due´ to low motivation, cognitive overload, or diminishing interest. When human learners constitute the environment, disengagement is inherently stochastic rather than deterministic. A learner's engagement may gradually fade, with dropout potentially occurring at any moment. There is no definitive terminal state that unequivocally signals the end of learning. To better reflect the dynamics of online learning, we replace the fixed discount factor $\gamma$ with a time-varying estimate of learner engagement probability, thereby accommodating dynamic uncertainty in a learner's future participation. Specifically, we introduce $p_t^{\text{eng}}$, the probability that a learner will remain engaged following the $t$-th learning interaction. This estimate, computed before the $t$-th step based on the learner's state $s_t$ and recommended item $a_t$, serves as a proxy for the learner's engagement level. Intuitively, assigning overly difficult content may offer high learning gains but increase the risk of dropout, while recommending only familiar material may fail to challenge the learner, leading to boredom. The value $p_t^{\text{eng}}$ captures this trade-off, reflecting the likelihood that the learner persists given state $s_t$ and action $a_t$. Accordingly, we revise the classical Q-function by incorporating dynamic $p_t^{\text{eng}}$ in place of the fixed discount factor $\gamma$. The modified Q-function is given by:

$$Q_\pi(s_t, a_t) = r_t + p_t^{\text{eng}} \mathbb{E}_{s_{t+1}, a_{t+1} \sim \pi} [Q_\pi(s_{t+1}, a_{t+1})], \quad (7)$$

Now, the action $a_t$ at the $t$-th interaction influences not only the immediate reward $r_t$, but also the learner's likelihood of continued engagement, denoted by $p_t^{\text{eng}}$. This engagement probability, in turn, affects the expected reward of future learning interactions, starting from step $t+1$. By incorporating time-varying $p_t^{\text{eng}}$ into the Bellman equation, the adjusted Q-function accounts for the stochastic nature of learner disengagement, yielding more realistic value estimates. This engagement-aware formulation allows the RL agent to function both as a risk manager—recognizing when future rewards may be lost due to disengagement—and as a motivational coach—identifying opportunities to challenge and retain committed learners. As a result, the AI-Tutor can learn policies that optimize both academic progress and sustained learner motivation.

We approximate $Q_\pi$ with a parameterized critic network $Q_\theta$, which is trained by minimizing the Bellman residual:

$$\mathcal{L}_{\text{critic}}(\theta) = \mathbb{E}_{(s_t, a_t, r_t, s_{t+1}) \sim D} \left( Q_\theta(s_t, a_t) - y_t \right)^2, \quad (8)$$

where $D$ is the replay buffer containing historical interaction tuples $(s_t, a_t, r_t, s_{t+1})$. The use of a replay buffer helps stabilize training by breaking temporal correlations and enabling efficient reuse

of past experiences. The target value $y_t$ is computed as:

$$y_t = r_t + p_t^{\text{eng}} \mathbb{E}_{a_{t+1} \sim \pi_\phi} Q_{\theta'}(s_{t+1}, a_{t+1}), \quad (9)$$

where $Q_{\theta'}$ is a slowly updated target network that stabilizes training, and $\pi_\phi$ is a parameterized actor network with parameters $\phi$. The actor network $\pi_\phi$ is trained to select actions that maximize expected long-term rewards, based on feedback from the critic. The training objective is:

$$L_{actor}(\phi) = E_{s_t \sim D, a_t \sim \pi_\phi}[Q_\theta(s_t, a_t)],\tag{10}$$

which encourages the actor to favor actions that yield higher value under the critic's evaluation.

Inspired by Haarnoja et al. (2018), we incorporate an entropy regularization term into the Qfunctiontopromoteexploratorybehavior.ThedetailsarelistedinAppendixC.4.Thisregularization enhances the diversity of the learned policy, improving robustness and stability during learning.

### 3.4. Modeling Learner Behavior: Environment Simulator for Model-Based RL

With the above setup, the RL model could interact directly with the learner to continuously collect transition tuples $\{s_t, a_t, r_t, s_{t+1}\}$, which are stored in a replay buffer $D$. The data in this buffer is then used to update both the critic and actor networks by optimizing EQ 8 and 10. This approach is generally known as a model-free RL training scheme. However, training an RL via a model-free scheme typically requires a large volume of interaction data to sufficiently explore diverse action trajectories and iteratively refine the policy. This introduces practical risks, such as deploying untested or suboptimal policies directly with learners may degrade learner experience, and also raises concerns regarding data inefficiency (Moerland et al. 2023). To address these challenges, researchers adopt model-based reinforcement learning, which incorporates an environment simulator to approximate the environment's responses under various state-action pairs (Kokkodis and Ipeirotis 2021, Wang et al. 2023). Model-based approaches are widely recognized for their superior sample efficiency compared to model-free methods and have been successfully applied in a range of RL applications. In this study, we propose a Learner Simulator (LearnSim) to model learner behaviors. Specifically, for each incoming $t$-th learning interaction, LearnSim estimates the key quantities required for training the AI-Tutor: the engagement probability $p_t^{\text{eng}}$, the recall probability $p_t^{\text{rec}}$, and the half-life strengths $h_{pre(t)}$ and $h_{post(t)}$. These estimates allow us to fully estimate entities in the reward function defined in EQ 2, and to incorporate $p_t^{\text{eng}}$ into the revised Q-function in EQ 9, thereby generating complete tuples to populate $D$. This closes the loop for training the RL in a model-based manner.

Similar to learner state representation in Section 3.2, we represent each learner through their learning trajectory, which summarizes the learner's evolving knowledge state over time. The trajectory up to time $t$ is represented as a sequence $L_t = \{\mathbf{l}_1, \ldots, \mathbf{l}_t\}$. Given the sequential nature of $L_t$ and the need to generate accurate predictions for each upcoming learning interaction, we adopt the Transformer architecture (Vaswani et al. 2017) to process the input sequence $L_t$ and generate the predictions. The Transformer is particularly well-suited for this task for several key reasons. First, its self-attention mechanism allows the model to capture long-range dependencies across the learning sequence. This is critical in educational contexts where the influence of prior interactions is not limited to adjacent time steps but may span the entire history of learning. Second, the Transformer's ability to dynamically assign attention weights allows it to identify and focus on the most relevant parts of the learning trajectory. This flexibility is important because the pedagogical significance of a knowledge item often depends on its context—e.g., its role as a prerequisite or its semantic proximity to the current topic. Third, the Transformer is highly scalable and computationally efficient, particularly during training and inference. As illustrated in Figure 4, we employ a Transformer model to process the student's learning trajectory and generate predictions for each incoming interaction. Specifically, for the $t$-th interaction, the attention memory processed by the Transformer model is denoted as $\boldsymbol{M}_t$, which is used to to predict the following quantities: 1) $p_t^{\text{eng}}$: the probability that the learner will continue engaging after the $t$-th interaction. 2) $p_t^{\text{rec}}$: the probability that the learner can recall the knowledge item prior to the $t$-th interaction. 3) $h_{post(t)}$: the half-life strengths of the learned knowledge item $a_t$ after the $t$-th interaction.

**Figure 4** **Framework of LearnSim for Multi-Task Learner Behavior Prediction**

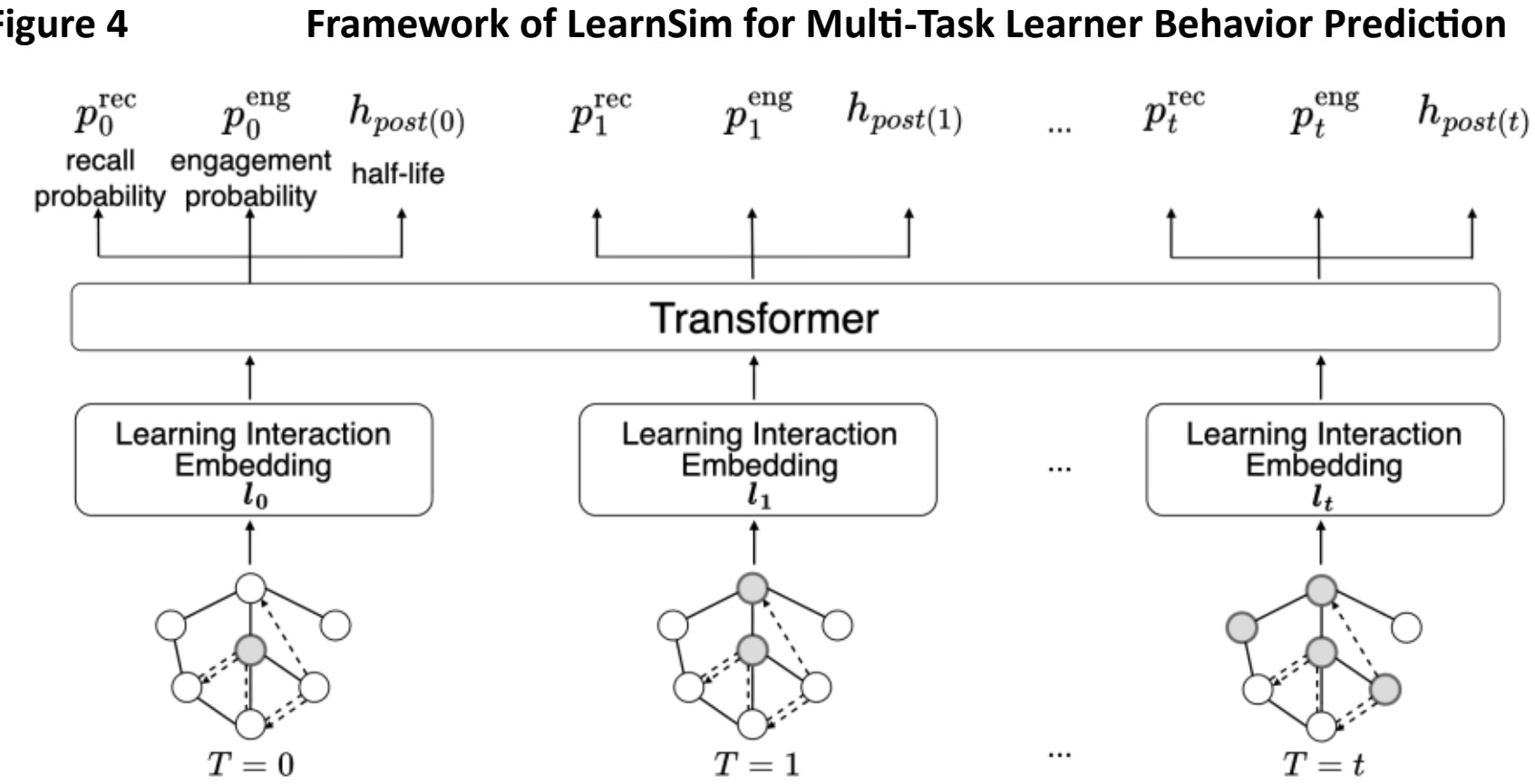

The first two predictions are probabilities and are generated via two separate neural networks with sigmoid activation on $\boldsymbol{M}_t$. The ground truth labels for these tasks can be directly obtained from historical learning data. For the first prediction, the target $\text{eng}_{t+1} \in \{0,1\}$ indicates whether the learner continues after the $t$-th interaction. For the second prediction, $\text{rec}_t \in \{0,1\}$ captures whether the student recalls the knowledge item before the learning in the $t$-th interaction[1]. If the student already knows the content, the system briefly reviews the key points; otherwise, they engage in full learning of the knowledge item. We employ standard binary cross-entropy loss functions to train

these two prediction heads. The loss functions are defined as:

$$\text{Loss}_{\text{eng}} = -\sum_t \text{eng}_{t+1} \cdot \log(p_t^{\text{eng}}) + (1-\text{eng}_{t+1}) \cdot \log(1-p_t^{\text{eng}}), \tag{11}$$

$$\text{Loss}_{\text{rec}} = -\sum_t \text{rec}_t \cdot \log(p_t^{\text{rec}}) + (1-\text{rec}_t) \cdot \log(1-p_t^{\text{rec}}). \tag{12}$$

For the third prediction—the half-life strength $h_{post(t)}$ of a knowledge item after the $t$-th learning interaction—the modeling is less straightforward than for the first two components. We still apply a neural network to the attention memory $\boldsymbol{M}_t$ to predict $h_{post(t)}$, where the output layer is equipped with a ReLU activation function to ensure non-negativity of $h_{post(t)}$. To supervise this prediction, we track the learner's next interaction with the same knowledge item at a future time step $t' > t$, and use the observed recall outcome $\text{rec}_{t'} \in \{0,1\}$ as the ground truth indicating whether the student successfully recalls the knowledge item. Given the time lag $\Delta t = t' - t$ and the predicted half-life strength $h_{post(t)}$, we estimate the recall probability using the half-life model introduced in EQ 4: $p_{t'}^{\text{rec}} = 2^{-\Delta t / h_{post(t)}}$. This predicted recall probability can be used to define a cross-entropy loss for training the model to accurately predict $h_{post(t)}$:

$$\text{Loss}_{\text{ret}} = -\sum_t \text{rec}_{t'} \cdot \log(p_{t'}^{\text{rec}}) + (1-\text{rec}_{t'}) \cdot \log(1-p_{t'}^{\text{rec}}). \tag{13}$$

[1] In our empirical setting of English vocabulary learning, the system prompts the learner during each interaction to indicate whether they can recall a particular word. Similar recall-check mechanisms are commonly used with other subjects (e.g., law and medicine) on major online learning platforms such as Quizlet (Quizlet 2024), Khan Academy (Khan 2025), and Duolingo (Duolingo 2023).

Using a similar approach, we can also estimate the prior half-life strength $h_{pre(t)}$ before the $t$-th interaction. This allows us to compute the knowledge retention reward $r_t^{\text{ret}} = h_{post(t)} - h_{pre(t)}$ as defined in EQ 5. Combining the three loss functions defined in EQ 11, 12, and 13, we train the LearnSim model on historical learning data by minimizing the total loss:

$$L_{\text{LearnSim}} = \text{Loss}_{\text{eng}} + \lambda_1 \text{Loss}_{\text{rec}} + \lambda_2 \text{Loss}_{\text{ret}}, \tag{14}$$

where $\lambda_1, \lambda_2$ are hyperparameters and will be tuned via grid-search to optimize the performance.

### 3.5. End-to-End Training of AI-Tutor

With the RL model and LearnSim in place, we now present the unified framework that integrates these components to learn an optimal recommendation policy. As illustrated in Figure 5, the overall architecture operates in an end-to-end manner, where the RL agent interacts with LearnSim to continuously refine its recommendation strategy. The detailed training procedure is summarized in Algorithm 1. Once LearnSim is trained using historical learning records, it serves as an environment simulator within the RL training loop. The RL training process consists of two main components: 1) Interaction and Data Collection: The actor interacts with the LearnSim environment to simulate

learningepisodes.Ineachepisode,theagentcollectstransitiontuples $(s_t, a_t, r_t, s_{t+1})$ andstoresthem intoreplaybuffer $\mathcal{D}$.ThisprocesscorrespondstoLines4-12inAlgorithm1.2)PolicyOptimization: Mini-batches are sampled from buffer $\mathcal{D}$ to update the actor and critic networks. Specifically, the critic network is updated by minimizing the Bellman residual loss (EQ 8), while the actor is

**Figure 5** **End-to-End Training Framework of the AI-Tutor**

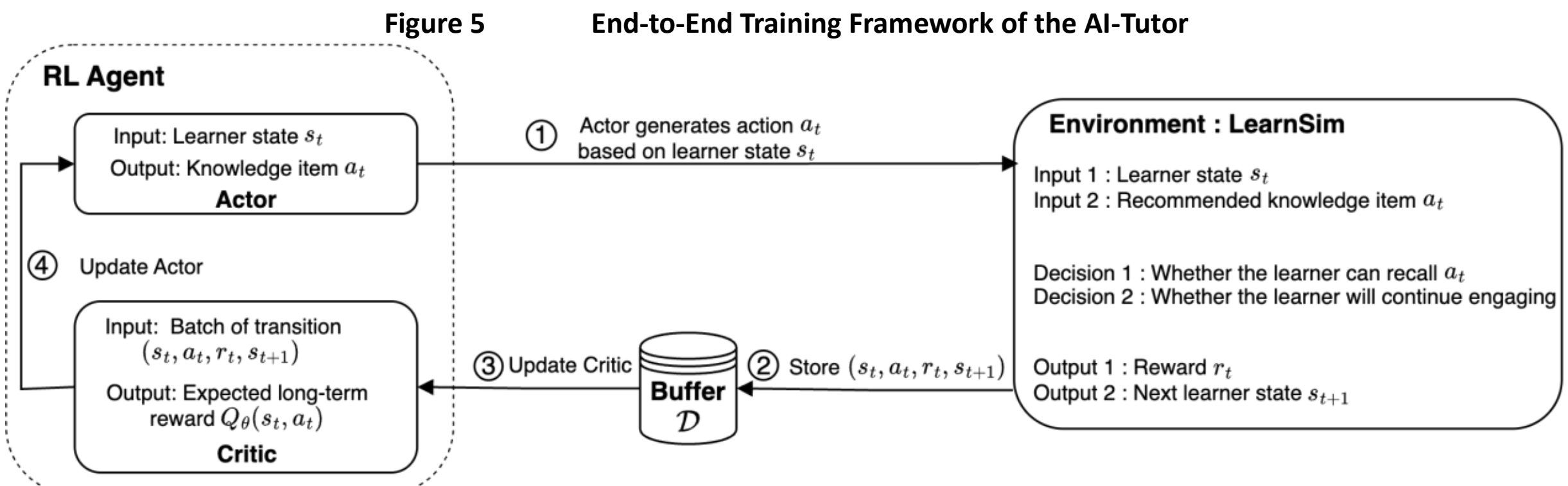

updated by maximizing the estimated long-term reward (EQ 10). This corresponds to Lines 13-17 in Algorithm 1. These two components are executed iteratively within the training loop (Lines 3-18), allowing the agent to progressively improve its policy through simulated feedback. By combining the learner behavior dynamics modeled by LearnSim with the policy refinement capabilities of the AC framework, AI-Tutor enables effective and sustainable learning path optimization in an end-to-end manner. To summarize, we highlight three key innovations of the proposed AI-Tutor:

**Algorithm 1** Training AI-Tutor via Model-based Reinforcement Learning

1: **Input:** Trained LearnSim.
2: **Initialize:** Actor policy $\pi_\phi$, critic network $Q_\theta$, replay buffer $D$.
3: **for** each episode **do**
4:     **while** eng$_t$ == 1 and course is incomplete **do**
5:         Observe current state $s_t$ and feasible action space $A_t$ from knowledge graph $G$.
6:         Select an action based on the current policy: $a_t \sim \pi_{a_t \in A_t}(a_t \mid s_t^{\prime\phi})$.
7:         LearnSim makes predictions of $p^{\text{eng}}{}_t$, $p^{\text{rec}}{}_t$, and $h_{post(t)}$.
8:         Observe the reward $r_t$ via EQ 2, next state $s_{t+1}$.
9:         Store the full transition ($s_t, a_t, r_t, s_{t+1}$) in the replay buffer $D$.
10:         Sample the next engagement eng$_{t+1} \sim p^{\text{eng}}{}_t$ .
11:         Update $t = t + 1$
12:     **end while**
13:     **for** each gradient update step **do**
14:         Sample a mini-batch of transitions from $D$.
15:         Update critic $\theta$ by minimizing the TD error loss $L_{\text{critic}}$ (EQ 8).
16:         Update actor $\phi$ using the policy gradient to maximize value $L_{\text{actor}}$ (EQ 10).
17:     **end for**
18: **end for**

1. **Knowledge-Graph-Based Representation and Action Guidance:** A KG is constructed to structurally organize the knowledge items. This graph serves multiple purposes. First, it enables dynamic representation of each learner's knowledge state by tracking their learning trajectory. Second, it provides structural representations of knowledge items, capturing both their semantic and relational properties. Third, the graph supports action masking to recommend pedagogically appropriate items, following the constructivist learning theory (Von Glasersfeld 2012), which posits that learning is most effective when content is slightly beyond the learner's knowledge boundary.

2. **Cognitive-Theory-Driven RL for Balancing Competing Goals:** The RL component of AITutor is designed to balance two critical trade-offs: (1) engagement versus learning progress, and (2) new knowledge acquisition versus long-term retention. To operationalize these goals, we customize the reward function to incorporate both immediate learning gains and contributions to long-term memoryretention,groundedincognitivepsychologytheoriesofmemorization(Finkenbinder1913, Ebbinghaus 2013, Radvansky et al. 2022). Moreover, we revise the classical Bellman equation by incorporating a dynamic estimate of learner engagement probability into the long-term value calculation. These advancements enable the RL to learn policies that not only sustain motivation but also optimize for both knowledge expansion and durable knowledge retention.

3. **Model-Based RL via Learner Simulator:** To support efficient and effective policy training, the RL agent interacts with LearnSim built on a Transformer model. LearnSim is designed for multitask learning and can reliably predict key learner responses, including engagement, knowledge recall, and memory strength. By accurately modeling learner behavior, LearnSim enables modelbased RL training, which significantly enhances data efficiency and improves training stability.

# 4. Empirical Evaluation

## 4.1. Context and Data

We collaborate with *MaiMemo*[2], one of the largest online language-learning platforms in Asia, for this study. MaiMemo primarily focuses on English vocabulary learning and serves over 2 million users daily. Upon selecting a course (e.g., GRE vocabulary), learners specify a daily goal for the number of words to study. During each learning interaction, the app presents a word and asks whether the learner has already mastered or can recall it. If the learner indicates unfamiliarity, the system provides comprehensive learning materials, including definitions and usage examples. If the learner reports prior mastery, the app delivers a brief review of the core content to reinforce memory. As with most online education platforms, learners can pause or drop out at any time. A detailed description of the user learning journey and the app interface is provided in Appendix B.

**Table 1 Data Summary**

| Category | Metric | Value |
|---|---|---|
| Platform Dataset Size | Total learners Number | 33,700 |
| | Total Learning Interactions | 23,656,968 |
| Knowledge Graph | Total KG Nodes (Total Vocabulary Words) | 6,180 |
| | Total KG Edges | 19,096,200 |

[2] https://www.maimemo.com

| | | |
|---|---|---|
| User Behavior Pattern | Recall Probability Avg | 0.705 |
| | Recall Probability Std | 0.456 |
| | Learning Sessions Per Person Avg | 71.987 |
| | Learning Sessions Per Person Std | 41.184 |
| | Learning Sessions Per Person Median | 14 |

To ensure that our analysis captures naturally occurring and unbiased learning behaviors, we use data from learners interacting with MaiMemo's randomized word recommendation system, rather than from its proprietary adaptive algorithms, which may restrict exploration or introduce estimation biases. The final dataset comprises over 23 million learning records from 33,700 randomly selected learners studying more than 6,000 English vocabulary words for the Graduate Entrance Examination. The data span from Jan 2022 to Sep 2023, providing a broad temporal window to observe diverse learning trajectories and long-term outcomes. Summaries are provided in Table 1.

We used 50% of the learners' fine-grained study records as training data to develop the LearnSim (Martens et al. 2016) and allocated 10% as a test set to evaluate its performance. Since the AI-Tutor adopts a model-based RL schema, its training involves interacting exclusively with the trained LearnSim. However, during policy evaluation, the model typically interacts with a different environment than the one used for training. Following the standard practice in the literature (Kokkodis and Ipeirotis 2021, Wang et al. 2023), we used the remaining 40% of the data to train a separate LearnSim, which serves as the testing environment to evaluate the AI-Tutor. Details on knowledge graph construction, model training, and hyperparameter tuning are provided in Appendix C.

### 4.2. Learner Simulator Evaluation

The LearnSim generates three predictions for each interaction: $p_t^{\text{rec}}$, the probability that the learner can successfully recall the recommended knowledge item $a_t$; $p_t^{\text{eng}}$, the probability that the learner will continue engaging after the $t$-th interaction; and the half-life of memory strength $h_{post(t)}$ for $a_t$ following the $t$-th interaction. The first two are straightforward binary classification tasks. The third, as described in Section 3.4, is used to estimate the future recall probability of the same item $a_t$ and is therefore also treated as a binary classification problem. Thus, to evaluate the simulator's predictive performance, we use accuracy, F1 score, AUC, and average precision. Because the engagementpredictiontaskisimbalanced,withmostobservationsreflectingcontinuedengagement, we additionally report balanced accuracy—the arithmetic mean of sensitivity and specificity—as it is well-suited for evaluating classifiers under class imbalance.

We first evaluate two related tasks: immediate-next knowledge recall prediction using $p_t^{\text{rec}}$ and future knowledge recall prediction using the predicted half-life $h_{post(t)}$ of each interaction. Due to the similarity between these two tasks, we adopt a shared set of benchmark models for comparison:

- *HLR (Half-Life Regression)* (Settles and Meeder 2016): A widely adopted model in industry that estimates the exponential decay of memory for each knowledge item, equivalent to EQ 4. It also serves as the organic memory model developed by Duolingo.
- *DHP-HLR* (Ye et al. 2022): This model extends HLR by incorporating a Markov property, which decomposes the time-series learning history into discrete state variables, offering a more structured view of learning dynamics.
- *GRU-HLR* (Su et al. 2023): This approach replaces manually specified state transitions with a Gated Recurrent Unit (GRU) network, enabling the model to automatically learn the memory state updates between review events.
- *LearnSim$_{v1}$*: An ablation variant of our proposed model, LearnSim$_{v1}$ replaces word representation from KG embeddings to widely used Word2Vec embeddings (Mikolov et al. 2013), allowing us to isolate the contribution of structured knowledge representations from KG.

Figure 6 compares model performance across key metrics. For immediate next-item knowledge recall prediction, LearnSim consistently outperforms baseline models, achieving at least a 33.39% improvement in AUC and a 36.74% improvement in balanced accuracy. The performance gain over LearnSim$_{v1}$ further highlights the benefit of incorporating KG, which contributes an additional 9.39% increase in AUC. Similar performance patterns are observed in the evaluation of future knowledge recall prediction, reaffirming the robustness of the proposed modeling approach.

To assess LearnSim's effectiveness in predicting learner engagement, we benchmarked LearnSim against three widely used approaches in engagement/churn prediction: logistic regression, random forest, and gradient boosting (Martens et al. 2016, Lemmens and Gupta 2020). These models represent a range of paradigms with varying levels of interpretability, complexity, and predictive performance, providing a well-rounded basis for comparison. As shown in the Figure 6, LearnSim

**Figure 6** **Performance Comparison of Different Models on Learner Behavior Predictions**

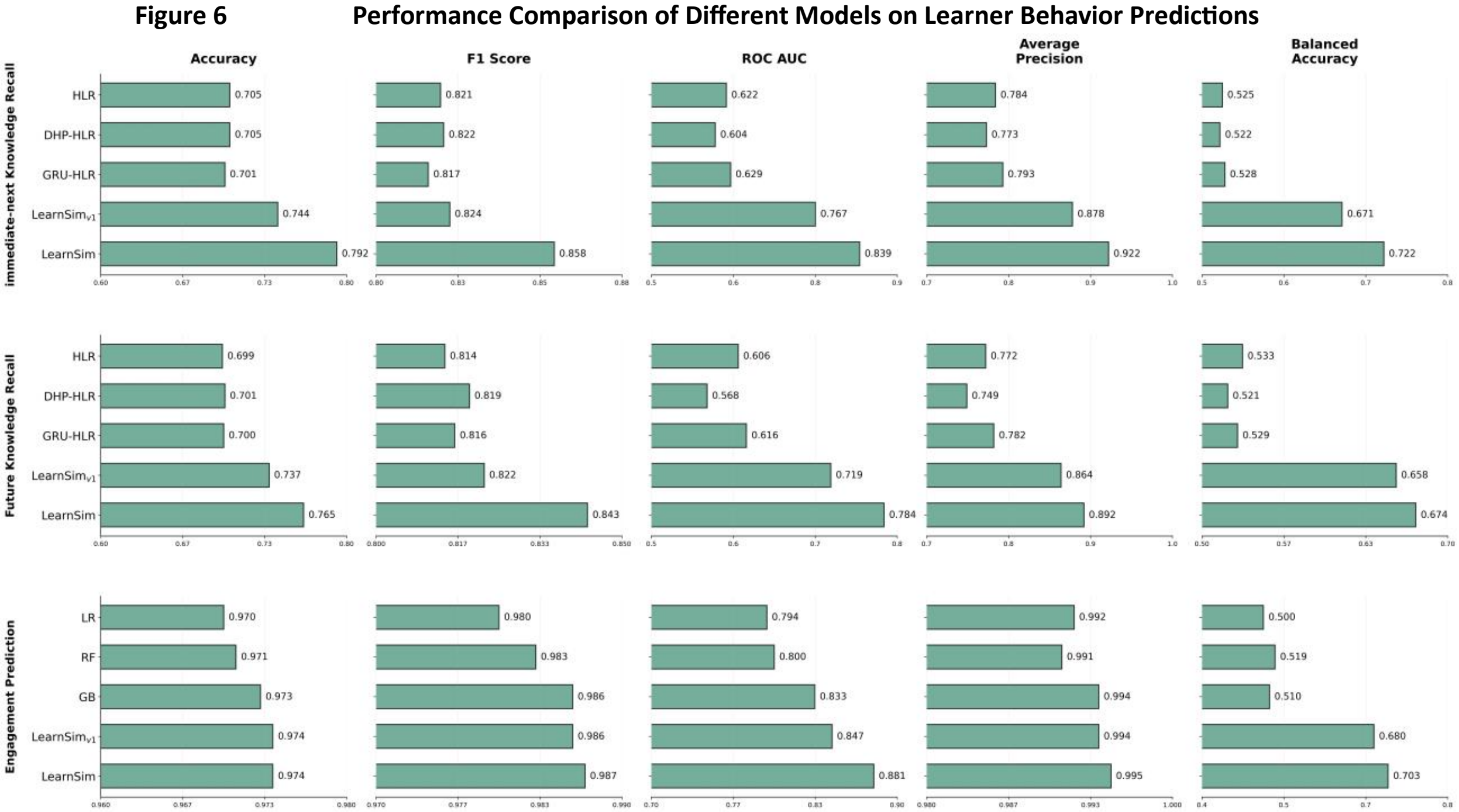

*Note.* LR : Logistic Regression, RF : Random Forest, GB : Gradient Boosting. Given the large dataset size, performance metrics on the held-out test set are stable, rendering confidence intervals from computationally expensive cross-validation unnecessary.

again demonstrates a clear advantage, improving AUC by a margin of 5.76% to 10.97% and dramatically increasing balanced accuracy by over 35% against all three models.

The superior performance of LearnSim across all tasks stems from two core design choices. First, its architecture integrates a Transformer-based attention mechanism with knowledge graph embeddings, allowing it to effectively capture sequential behavior and semantic relationships between concepts.Second,asamulti-taskmodel,LearnSimjointlylearnsfromknowledgerecallandengagement prediction, enabling richer learner representations than single-task models. These results validate LearnSim as a robust and reliable environment simulator for downstream RL applications.

### 4.3. AI-Tutor Evaluation

To evaluate the performance of the AI-Tutor, we set up the following English vocabulary learning scenario. A total of 1,000 learners attempt to master 1,000 English vocabulary words over a 28-day cycle—a typical learning horizon for online English learning platforms. Each day, every learner engages in 150 learning interactions with the AI-Tutor. As specified in Section 4.1, a task-specific instance of LearnSim is trained as the testing environment for this evaluation. Mirroring realworld conditions, LearnSim allows learners to discontinue the course at any point based on their predicted engagement probabilities. The initial states of the 1,000 learners are randomly sampled from real-world study records on *MaiMemo* platform. We track each learner's progress throughout the learning period by interacting with AI-Tutor, capturing both their performance and potential dropout behavior. At the end of the 28-day course, we evaluate two key outcomes: (1) Course completion rate, defined as the proportion of learners who persist through the entire learning period; and (2) Final exam performance, assessed based on a test covering all 1,000 vocabulary words for all 1,000 learners. Each student's response to the final test is sampled by the estimated recall probability of each word based on the learner's memory strength (i.e., half-life). These exam results offer a comprehensive assessment of long-term learning outcomes, both across the entire cohort and within the subgroup of course completers. To ensure robustness, we repeat this evaluation procedure 100 times and report the resulting estimates and distributions. We compare the performance of AI-Tutor with following benchmarks and its ablation variants under the same experimental setup:

- *Myopic Greedy*: A purely exploitative baseline that always selects the item with the lowest current recall probability. Consequently, the model prioritizes new and difficult items.
- *FSRS$_{v1}$*: FSRS (2024) released a widely adopted open-source system for spaced repetition. FSRS recommends items for review when their estimated retention value falls below a predefined threshold (e.g., 0.5). Following standard industry practice, we use Half-Life Regression (HLR) (Settles and Meeder 2016) to estimate the memory strength of each knowledge item. This approach is widely used by platforms such as MaiMemo and Anki[3].
- *FSRS$_{v2}$*: This variant adopts the same scheduling logic as FSRS$_{v1}$ but replaces HLR with the LearnSim to estimate memory strength, leveraging its superior prediction performance on this task.
- *DRL-SRS*: Xiao and Wang (2024) propose a deep Q-learning framework for optimizing personalized spaced repetition schedules. Their model treats memory retention as a dynamic process and adaptively selects review items to maximize long-term recall probability.
- *AI-Tutor$_{v1}$*: This ablation model of AI-Tutor excludes engagement awareness by replacing the churn probability $p_t^{eng}$ in the Q-function (EQ 7) with a constant discount factor $\gamma = 0.99$.

[3] https://apps.ankiweb.net/

- *AI-Tutor$_{v2}$*: This version *disables KG-based action masking*, allowing the agent to select from the full knowledge item space without structural guidance.

- *AI-Tutor$_{v3}$*: This model removes the knowledge retention component ($r_t^{\text{ret}}$) from the reward

function. The reward reduces to: $r(s_t, a_t) = r_t^{\text{acq}} = 1 - p_t^{\text{rec}}$.

Adetailedcomparisonofthemechanismsunderlyingallevaluatedmodelsispresentedin Table2, and their performance on course completion rates and final exam outcomes is summarized in

**Table 2** **Mechanism Comparison of Knowledge Recommendation Models**

| **Model** | Learner Engagement | Knowledge Retention | KG-based Action Masking | KG-based State Representation | RNN State Encoding | Reinforcement Learning |
|---|---|---|---|---|---|---|
| AI-Tutor | ✓ | ✓ | ✓ | ✓ | ✓ | ✓ |
| AI-Tutor$_{v1}$ | × | ✓ | ✓ | ✓ | ✓ | ✓ |
| AI-Tutor$_{v2}$ | ✓ | ✓ | × | ✓ | ✓ | ✓ |
| AI-Tutor$_{v3}$ | ✓ | × | ✓ | ✓ | ✓ | ✓ |
| DRL-SRS | × | × | × | × | ✓ | ✓ |
| FSRS$_{v1}$ | × | ✓ | × | × | × | × |
| FSRS$_{v2}$ | × | ✓ | × | ✓ | × | × |
| Myopic Greedy | × | × | × | × | × | × |

Notes: ✓ denotes the presence of the component, while × denotes its absence.

**Figure 7** **Performance Comparison of Knowledge Recommendation Models**

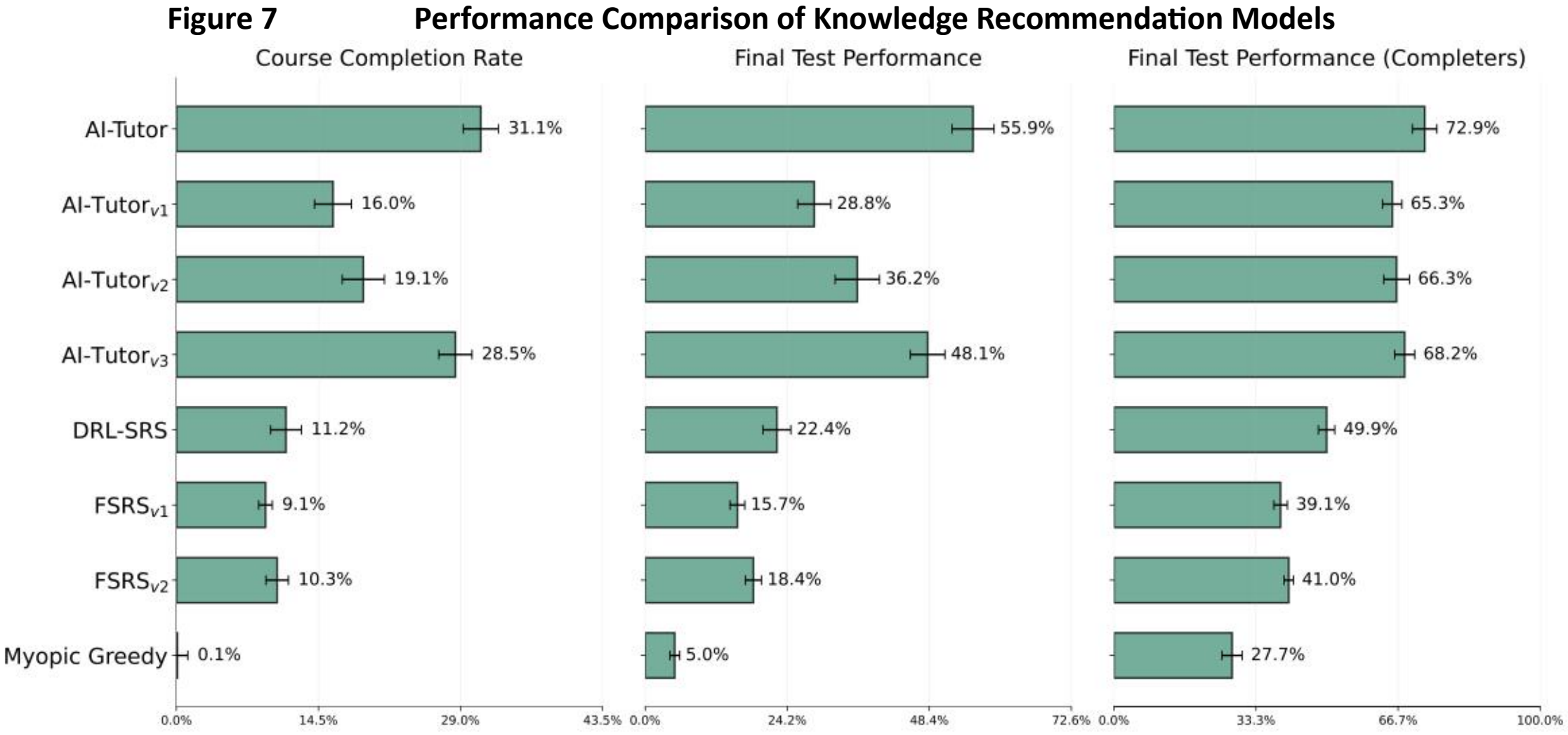

*Note.* The error bar represents the 95% confidence interval.

Figure 7. We observe that AI-Tutor significantly outperforms both the benchmark models and all ablation variants, demonstrating its superior ability to support sustainable learning. To further understand the dynamic behavior of learners over the 28-day cycle, we analyze two aspects: (1) the dynamics of learner engagement patterns, measured by sustained learner rates under different policies; and (2) the evolving knowledge coverage achieved by each policy, measured by the daily proportion of the 1,000 target words that have been recommended to the learner. Figure 8 presents these dynamics, from which we derive the following key insights:

1. Unsurprisingly, the myopic greedy method, a purely exploitative policy that always selects the item with the lowest current recall probability, performs the worst across all three evaluation metrics in Figure 7. As shown in Figure 8, this approach aggressively pushes learners to acquire new knowledge to maximize immediate knowledge acquisition. However, it neglects learner motivation

**Figure 8** **Learner Engagement Curve and Knowledge Coverage Curve**

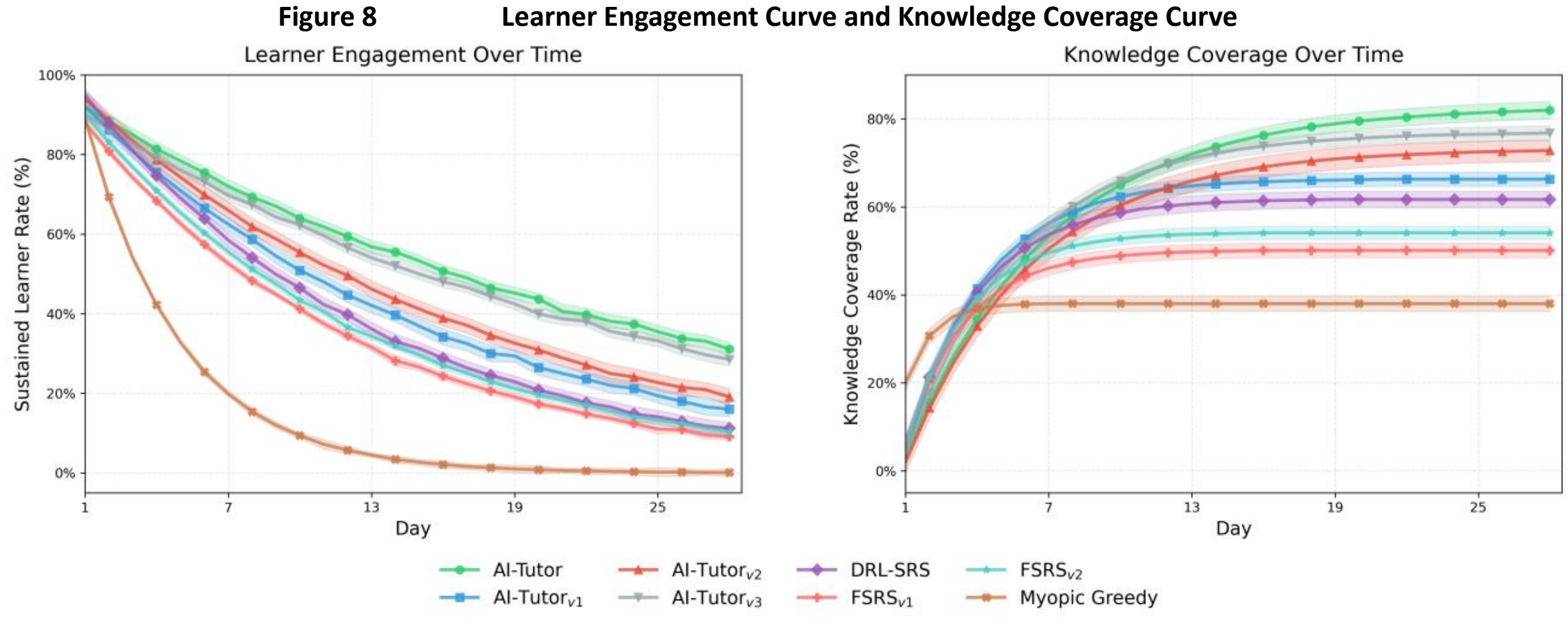

*Note.* The shaded bands represent the 95% confidence interval.

and cognitive limits, resulting in a sharp increase in dropout rates. By around day 20, nearly all users have disengaged. This outcome underscores the importance of balancing knowledge acquisition with sustained engagement; an overly aggressive strategy can ultimately backfire.

2. Both FSRS variants perform significantly better than the myopic greedy baseline, highlighting the advantages of memory-theoretic, rule-based approaches over na¨ıve heuristics. Among the two, FSRS$_{v2}$, which employs LearnSim to estimate long-term memory strength, outperforms FSRS$_{v1}$, which uses HLR. This comparison demonstrates that a more accurate estimation of knowledge retention leads to better review scheduling and improved learning outcomes.

3. While FSRS uses threshold-based rules to trigger reviews, DRL-SRS employs reinforcement learning to dynamically schedule review events. This RL-based approach systemically learns the adaptive policy and therefore achieves further improvements in long-term knowledge retention, showing the potential of reward-optimized strategies to outperform rule-based approaches.

4. All AI-Tutor ablation variants outperform the best non-ablation model, DRL-SRS. Notably, compared to DRL-SRS, AI-Tutor increases the course completion rate by 177.7% (from 11.2% to 31.1%). While the completion rate is still only 31.1%, it significantly surpasses all other models. AI-Tutor also improves final exam performance by 149.6% (from 22.4% to 55.9%), confirming the effectiveness of the proposed framework in advancing sustainable learning. By comparing the ablation variants, we have the following observations: First, removing engagement modeling ($AITutor_{v1}$) results in the most significant performance degradation, highlighting the critical role of explicitly incorporating engagement probability into the long-term reward estimation. Second, in $AI\text{-}Tutor_{v2}$, removing KG-based action masking leads to higher dropout rates without notable gains in knowledge acquisition; without this guidance, the RL agent struggles with an excessively large action space. Finally, although $AI\text{-}Tutor_{v3}$, which excludes the knowledge retention component in the reward function, performs the best among all ablations, it still lags significantly behind the full model in the final exam performance. This highlights the value of modeling long-term memory strength in promoting sustainable learning.

## 5. Learning Path Analysis

In the prior section, we quantitatively evaluated the performance of the AI-Tutor in knowledge recommendation. In the following section, we further explore the underlying mechanisms that distinguish the AI-Tutor from other models in guiding learning trajectories. We do this by examining specific learner paths across different benchmarks and conducting subgroup analyses. This approach allows us to uncover how the AI-Tutor dynamically adapts its personalized recommendation strategies to accommodate learners with diverse backgrounds and learning behaviors.

### 5.1. Impact of Engagement-Aware Policy on Learning Path

AI-Tutor explicitly incorporates engagement probability into the RL framework, enabling the system to optimize long-term learning outcomes while promoting a positive and sustainable learning experience. To evaluate the impact of this engagement-aware component on learning path guidance, we focus on $AI\text{-}Tutor_{v1}$, an ablation model of AI-Tutor that excludes engagement awareness. Additionally, since $FSRS_{v1}$ is one of the most widely adopted recommendation models in industry, we include it as another baseline. We randomly selected an initial learner state from *MaiMemo* records. Using this identical seed state, we simulate the learner's interactions with three different personalized tutors. The knowledge graph is visualized, with colored nodes indicating the learner's paths. The resulting learning paths generated by the three algorithms are illustrated over the KG in Figure 9. To facilitate interpretation, each node is color-coded based on word difficulty [4], providing visual cues about the cognitive load associated with each word. We observe several key differences in the learning trajectories produced by the three models:

1. Both AI-Tutor and AI-Tutor$_{v1}$ utilize KG-based action masking, which ensures that new learning items are selected from the adjacent boundary of the learner's current knowledge. As a result, their learning trajectories exhibit a clear expansion pattern along the edges of the mastered knowledge subgraph. In contrast, FSRS$_{v1}$ does not leverage any knowledge graph structure for learning the next item. Its selected items appear scattered across the graph, showing no coherent adjacency and connections to previously learned knowledge items.

**Figure 9** **Learning Pattern Comparison: With vs. Without Learner Engagement Modeling**

[4] Word difficulty is determined by *MaiMemo* based on learners' historical learning records and intrinsic word attributes.

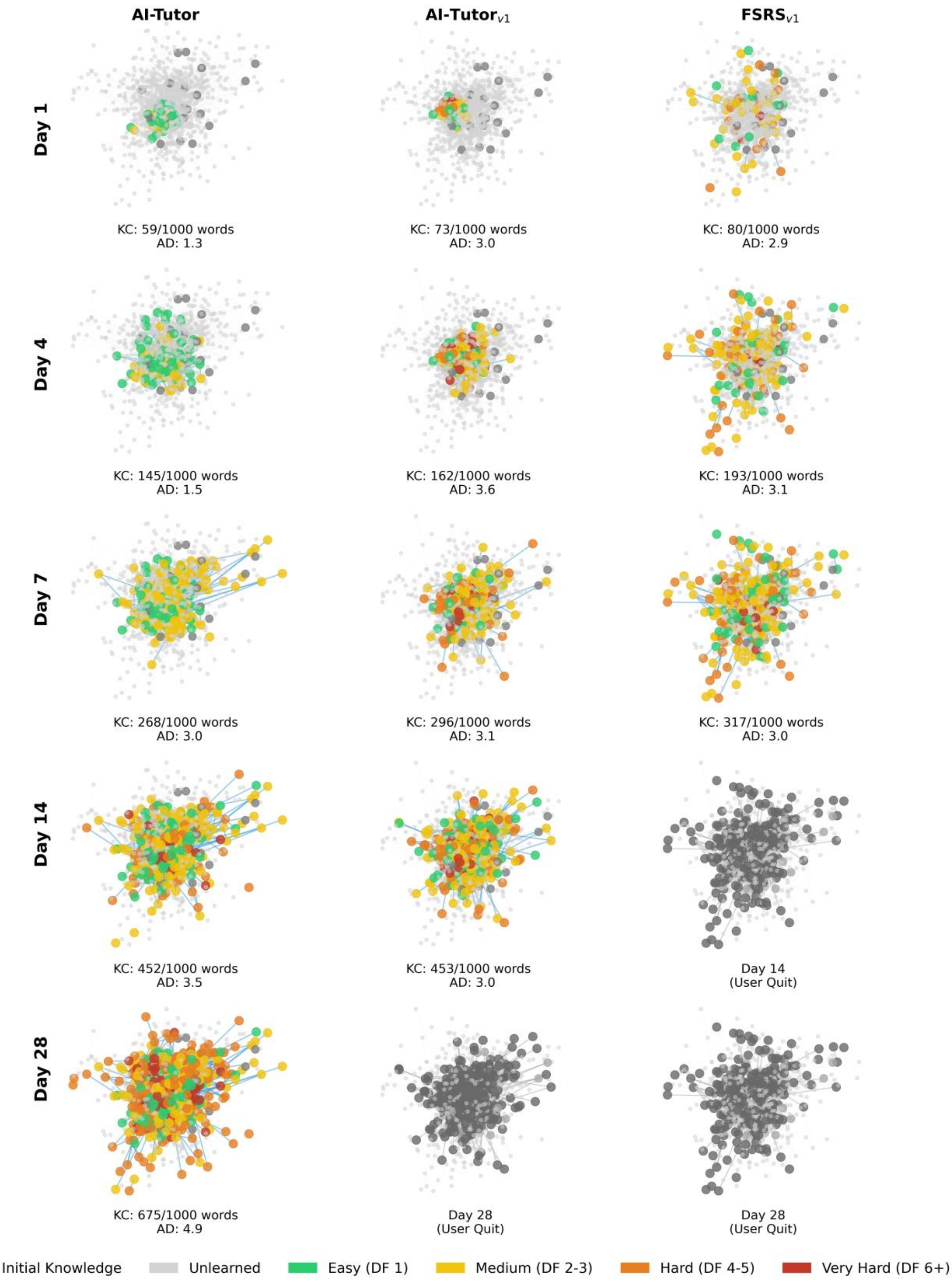

*Note.* This figure illustrates the learning progression of the same seed user with different models. Each dot in the graphs represents a knowledge item. Each snapshot in time includes two key metrics: "Knowledge Coverage (KC)," which is the total number of unique words learned out of 1000 till the given day, and "Average Difficulty (AD)," which refers to the average difficulty of words learned on the given day. The solid grey graphs shown indicate that those users had quit the study.

2. Neither $FSRS_{v1}$ nor AI-$Tutor_{v1}$ considers learner engagement in their recommendation policy. Consequently,bothmodelsadoptanaggressivestrategy—focusingexclusivelyonrewardviaknowledge acquisition and retention—without considering the cognitive constraints of human learners. As shown in the early stages of their learning paths, these models tend to overload students by introducing a large number of new and difficult words. Such aggressive strategies overwhelm the learner and lead to early dropout, undermining long-term learning success.

3. In contrast, AI-Tutor gradually increases the difficulty of recommended content, aiming to sustain learner motivation throughout the learning process. By initially reinforcing foundational knowledge and progressively introducing more difficult concepts, AI-Tutor fosters engagement and enables the learner to complete the full 28-day course. This reflects a more human-centric strategy that balances knowledge acquisition with learner well-being and engagement.

### 5.2. Impact of Knowledge-Retention-Aware Policy on Learning Path

AI-Tutor comprehensively considers both knowledge expansion and knowledge retention within its reward function. As a result, its learning policy balances the dual objectives of learning new knowledge and reviewing to reinforce knowledge retention. To examine the effect of this design, we compare AI-Tutor with two baselines: AI-$Tutor_{v3}$, an ablation variant that removes the knowledge retention component in the reward function and focuses solely on knowledge expansion, and $FSRS_{v1}$, the widely adopted industry standard. Following the approach in Section 5.1, we randomly select an initial learner state from *MaiMemo* records. Based on this seed state, we simulate the learner's interactions with each of the three tutors and visualize their resulting learning trajectories on the KG. These learning paths are illustrated in Figure 10, where each node is colored based on the number of review interactions a recommended word received on a given day. We observe several key differences in the learning paths generated by the three models:

1. $FSRS_{v1}$ adopts a rule-based review mechanism that triggers a review once the estimated long-term memory strength of an item falls below a predefined threshold. However, this method lacks a systematic strategy to proactively consolidate learned knowledge. As shown in Figure 10, the number of review interactions per day remains relatively stable, and the early stages are primarily devoted to introducing new content. This aggressive strategy overloads the learner with new information and leads to early dropout. Consequently, the learner fails to complete the course and achieves only 20.9 on the final exam, indicating poor long-term learning outcomes.

2. AI-Tutor$_{v3}$, which excludes the knowledge retention component from the reward function, also adopts a more aggressive learning strategy, focusing heavily on acquiring new knowledge for

**Figure 10** **Learning Pattern Comparison: With vs. Without Knowledge Retention Modeling**

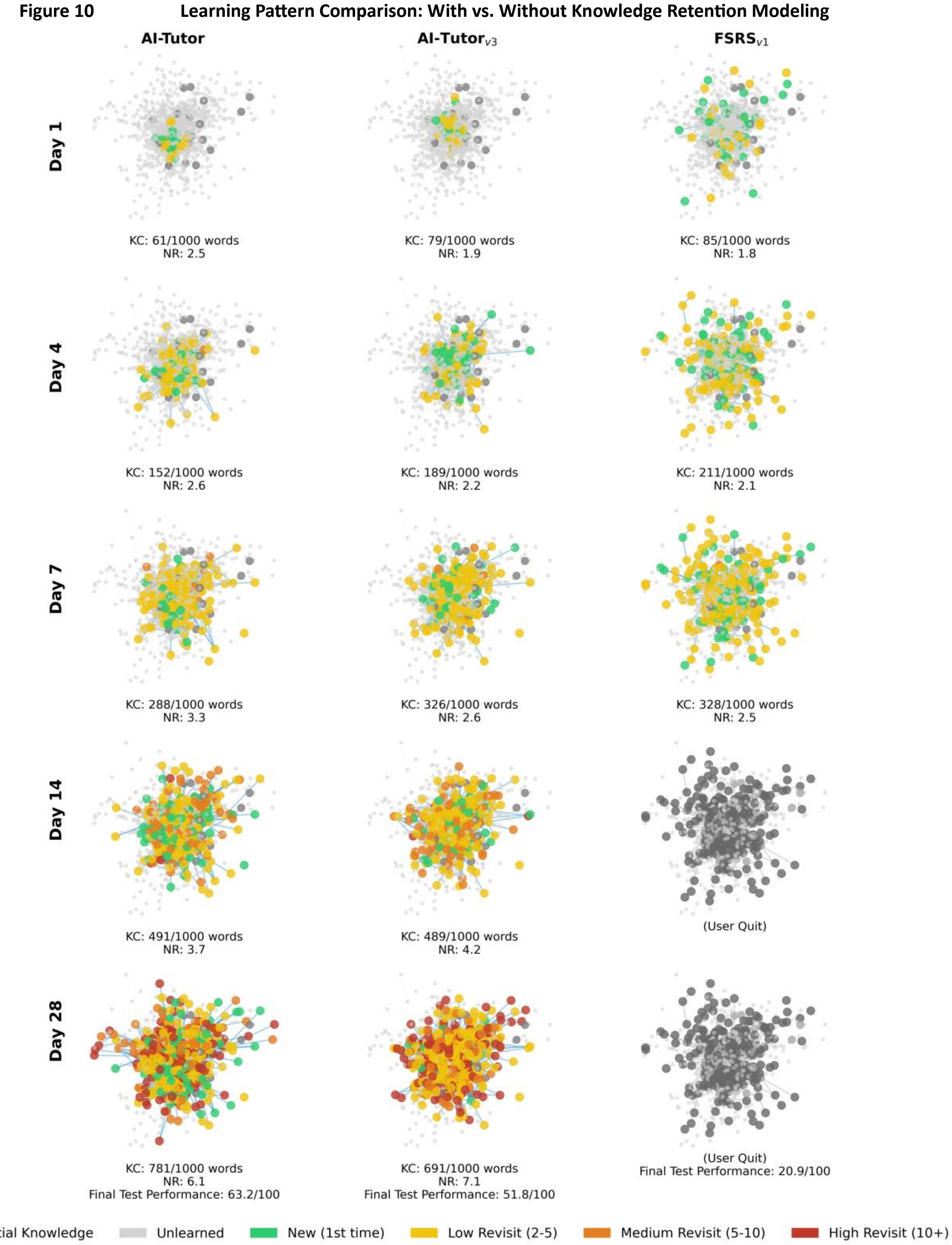

*Note.* This figure shows the learning progression of a seed user with different models. Each dot in the graphs represents a single knowledge item, with its color indicating its status. Each snapshot in time includes two key metrics: “Knowledge Coverage (KC),”

which is the total number of unique words learned out of 1,000 till the given day, and “Number of Reviews (NR),” which refers to the average number of review interactions a recommended word received on a given day. The final test score is shown at the end.

reward optimization. In the early stage, it covers significantly more vocabulary items than the full model. For example, by day 7, AI-Tutor$_{v3}$ has introduced 326 words, compared to 288 by AI-Tutor. However, due to the absence of a systematic review mechanism, many items are gradually forgotten after the mid-course period. As a result, the model must shift focus in the latter half of the course to compensate for the knowledge retention gap, spending substantial effort on reviewing. As illustrated in Figure 8, this strategy also leads to higher dropout rates in the later stages than the full model, reflecting a suboptimal learning experience.

3. AI-Tutor achieves a more balanced trade-off between knowledge expansion and knowledge retention. From the outset, it incorporates review to strengthen foundational memory, even though this leads to slightly fewer new items covered and a slower strategy in the early days. However, this “slow but steady” strategy pays off: by the end of the course, AI-Tutor has covered more total vocabulary items (781 vs. 691) than AI-Tutor$_{v3}$ and yields substantially higher final test scores (63.2 vs. 51.8), highlighting its effectiveness in promoting sustainable long-term learning.

### 5.3. Personalized Learning Path

To examine how AI-Tutor personalizes learning strategies to heterogeneous learners, we divide learners into three groups according to their performance in the initial state, which is randomly sampled from the empirical dataset. We define a learner’s initial performance as their average recall rate across the first 20 study interactions in the initial state, capturing the heterogeneous baseline before interacting with AI-Tutor. Based on this average recall rate, learners are categorized into three groups: Low-Performers (recall rate below 50%), Medium-Performers (between 50% and 80%), and High-Performers (above 80%). Each learner then engages with the AI-Tutor, and we analyze the learning paths and strategies deployed for each group. Specifically, we plot two learning trajectories: (1) the average difficulty level of learning materials assigned by AI-Tutor over time, and (2) the proportion of daily interactions devoted to reviewing previously learned content.

Figure 11 shows the learning trajectories for three groups. Overall, all three groups exhibit a general learning pattern of progressing from easier to more difficult materials. In terms of content, each group starts with a strong focus on acquiring new knowledge, while the proportion of review interactions increases steadily over time. Despite these commonalities, we observe the following key differences in learning strategies across the three groups:

1. For Low-Performers, the AI-Tutor adopts a conservative approach to content difficulty. Especially during the middle-to-late stages, the assigned materials consistently remain at a moderate to low difficulty level. In terms of review behavior, Low-Performers demonstrate a high proportion

**Figure 11** **Personalized Learning Patterns**

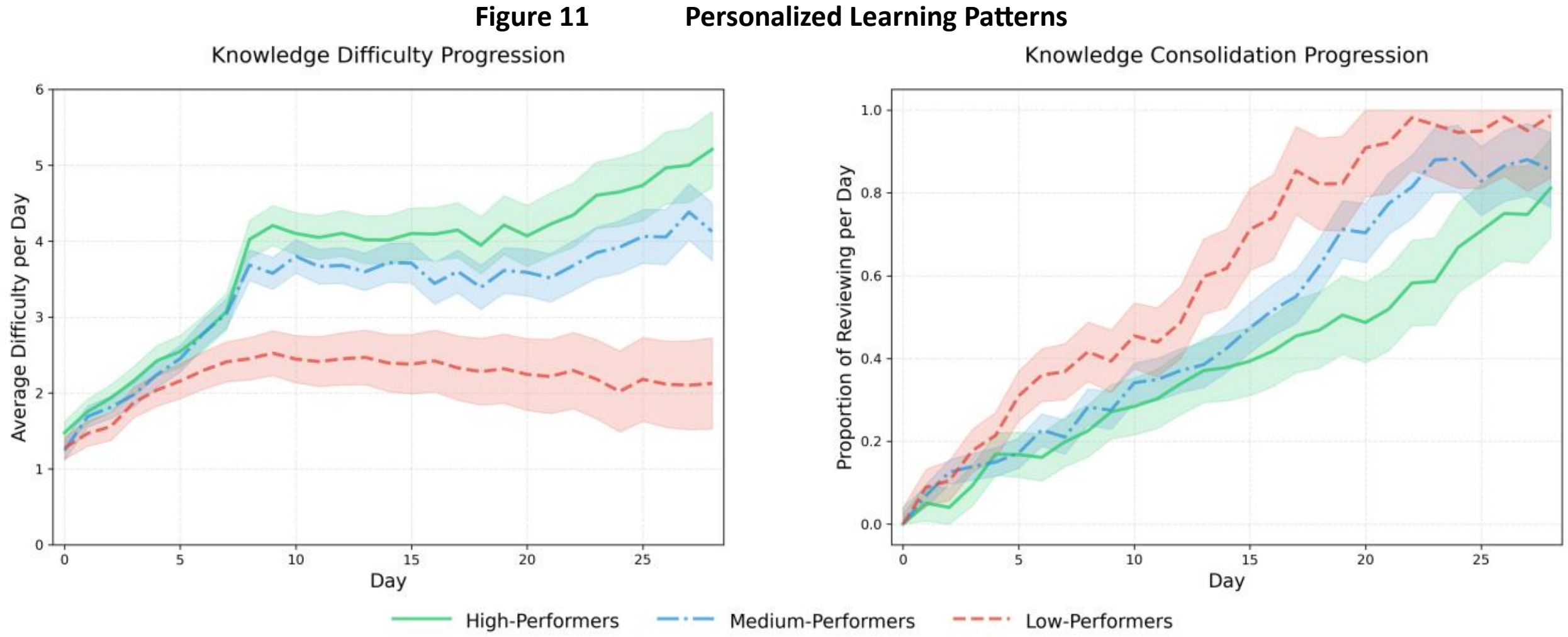

*Note.* The shaded bands represent the 95% confidence interval.

of review interactions from the outset, consistently surpassing the other two groups. This strategy suggests that the AI-Tutor effectively identifies their learning challenges and responds by assigning appropriately leveled content while emphasizing review to support long-term knowledge retention.

2. For Medium-Performers and High-Performers, content difficulty increases gradually in the early stage, stabilizes during the mid-phase, and rises again toward the end. The materials assigned to High-Performers are generally more challenging than those for Medium-Performers. In terms of review frequency, High-Performers require fewer review interactions, suggesting their more efficient learning and stronger long-term memory once content is acquired.

The learning path analysis reveals that AI-Tutor dynamically adapts its instructional strategies based on both learner engagement and long-term knowledge retention considerations. Compared to baseline models, it generates more coherent, human-centric learning trajectories that promote sustained engagement and long-term knowledge retention. Moreover, by personalizing difficulty levels and review intensity to match individual learner profiles, AI-Tutor demonstrates its capacity to support diverse educational needs. These insights can inform instructors' course design by providing data-driven guidance on sequencing and scaffolding (Wang et al. 2025).

## 6. Conclusion

In this paper, we present AI-Tutor, a reinforcement learning–based model designed to foster sustainable learning in online education. AI-Tutor leverages insights from cognitive theory to construct a reward function that balances the acquisition of new knowledge with reviewing learned knowledge to reinforce long-term memory. It further incorporates learner engagement into the RL training loop by modifying the Bellman equation to weight future cumulative rewards according to dynamically evolving engagement levels. These enhancements allow AI-Tutor to deliver personalized learning guidancethatachievesnotonlyshort-termlearningsuccessbutalsosustainedengagementandlongterm knowledge retention. Empirical evaluations on large-scale language learning data demonstrate that AI-Tutor consistently outperforms existing benchmarks and ablation variants across key metrics of learner engagement, knowledge retention, and overall learning outcomes. Through detailed trajectory analyses, we further show that AI-Tutor adapts its strategies to individual learner profiles, resulting in more effective and human-aligned learning experiences.

The core design of AI-Tutor lies in balancing two fundamental goals: knowledge expansion and knowledge retention, as well as learning progression and learner engagement. These dual objectives arecentraltonearlyallonlineeducationcontexts.Consequently,beyondlanguagelearning,AI-Tutor can be adapted to a wide range of subjects. For different disciplines, one only needs to construct a subject-specific knowledge graph based on the course syllabus and historical learning records. The subsequent process follows our framework: training the LearnSim model on the course's historical learning data, and then using LearnSim to train AI-Tutor. Many online education platforms and subject areas share patterns similar to those in our empirical setting: learners engage with course modules, undergo periodic reviews, and ultimately face a final exam. Given this general learning cycle, we believe AI-Tutor can be successfully extended to other domains. We leave this as an avenue for future research and practical exploration by other scholars and practitioners.

## References

Aggarwal CC, Li Y, Wang J, Wang J (2009) Frequent pattern mining with uncertain data. *Proceedings of the 15th ACM SIGKDD international conference on Knowledge discovery and data mining*, 29–38.

Altindag DT, Filiz ES, Tekin E (2024) Is online education working? *Educational Evaluation and Policy Analysis* 46(0):1–23.

Bassen J, Balaji B, Schaarschmidt M, Thille C, Painter J, Zimmaro D, Games A, Fast E, Mitchell JC (2020) Reinforcement learning for the adaptive scheduling of educational activities. *Proceedings of the 2020 CHI conference on human factors in computing systems*, 1–12.

Bauman K, Tuzhilin A (2018) Recommending remedial learning materials to students by filling their knowledge gaps. *MIS Quarterly* 42(1):313–A7.

Cai D, Zhang Y, Dai B (2019) Learning path recommendation based on knowledge tracing model and reinforcement learning. *2019 IEEE 5th international conference on computer and communications (ICCC)*, 1881–1885 (IEEE).

Cao J, Leng Y (2021) Adaptive data acquisition for personalized recommender systems with optimality guarantees on short-form video platforms. *Forthcoming, Management Science* .

Dempster FN (1989) Spacing effects and their implications for theory and practice. *Educational Psychology Review* 1:309–330.

Duolingo (2023) The duolingo method: How does duolingo teach new subjects? White paper.

Ebbinghaus H (2013) [image] memory: A contribution to experimental psychology. *Annals of neurosciences* 20(4):155.

Finkenbinder EO (1913) The curve of forgetting. *The American Journal of Psychology* 24(1):8–32.

FSRS (2024) Fsrs4anki: Free spaced repetition schedule for anki. URL https://github.com/ open-spaced-repetition/fsrs4anki/wiki/The-Algorithm.

Ghoshal A, Sarkar S (2014) Association rules for recommendations with multiple items. *INFORMS Journal on Computing* 26(3):433–448.

Grondman I, Busoniu L, Lopes GA, Babuska R (2012) A survey of actor-critic reinforcement learning: Standard and natural policy gradients. *IEEE Transactions on Systems, Man, and Cybernetics* 42(6):1291–1307.

Haarnoja T, Zhou A, Hartikainen K, Tucker G, Ha S, Tan J, Kumar V, Zhu H, Gupta A, Abbeel P, et al. (2018) Soft actor-critic algorithms and applications. *arXiv preprint arXiv:1812.05905* .

IMARC (2024) E-learning market report by imarc group. URL https://www.imarcgroup.com/ e-learning-market, accessed June 2025.

Jack R, Halloran C, Okun J, Oster E (2023) Pandemic schooling mode and student test scores: Evidence from us school districts. *American Economic Review: Insights* 5(2):173–190.

Kalashnikov D, Irpan A, Pastor P, Ibarz J, Herzog A, Jang E, Quillen D, Holly E, Kalakrishnan M, Vanhoucke V, Levine S (2018) QT-Opt: Scalable Deep Reinforcement Learning for Vision-Based Robotic Manipulation. *arXiv preprint arXiv:1806.10293*, available at https://arxiv.org/abs/1806.10293.

Khan (2025) Why khan academy will be using “skills to proficient” to measure learning outcomes.

Kokkodis M, Ipeirotis PG (2021) Demand-aware career path recommendations: A reinforcement learning approach. *Management science* 67(7):4362–4383.

Kumar A, Mehra A (2024) Improving educational delivery in k-12 schools with personalization: Evidence from a randomized field experiment in india. *Available at SSRN 2756059* .

Lemmens A, Gupta S (2020) Managing churn to maximize profits. *Marketing Science* 39(5):956–973.

Li AT, Liu D, Xu SX, Yi C (2024) Interleaved design for e-learning: Theory, design, and empirical findings. *MIS Quarterly* 48(4).

Liu Y, Yang Y, Chen X, Shen J, Zhang H, Yu Y (2020) Improving knowledge tracing via pre-training question embeddings. *arXiv preprint arXiv:2012.05031* .

Lockee BB (2021) Online education in the post-covid era. *Nature Electronics* 4(1):5–6.

Lyu S, Ling S, Guo K, Zhang H, Zhang K, Hong S, Ke Q, Gu J (2021) Graph neural network based vc investment success prediction. *arXiv preprint arXiv:2105.11537* .

Ma T, Hu Y, Lu Y, Bhattacharyya S (2024) Customer engagement prediction on social media: A graph neural network method. *Information Systems Research* .

Martens D, Provost F, Clark J, de Fortuny EJ (2016) Mining massive fine-grained behavior data to improve predictive analytics. *MIS quarterly* 40(4):869–888.

Mikolov T, Chen K, Corrado G, Dean J (2013) Efficient estimation of word representations in vector space. *arXiv preprint arXiv:1301.3781* .

Mnih V, Kavukcuoglu K, Silver D, Rusu AA, Veness J, Bellemare MG, Graves A, Riedmiller M, Fidjeland AK, Ostrovski G, et al. (2015) Human-level control through deep reinforcement learning. *Nature* 518(7540):529–533, URL http://dx.doi.org/10.1038/nature14236.

Moerland TM, Broekens J, Plaat A, Jonker CM, et al. (2023) Model-based reinforcement learning: A survey. *Foundations and Trends® in Machine Learning* 16(1).

Nakagawa H, Iwasawa Y, Matsuo Y (2019) Graph-based knowledge tracing: modeling student proficiency using graph neural network. *IEEE/WIC/aCM international conference on web intelligence*, 156–163.

Ning B, Lin FHT, Jaimungal S (2021) Double deep q-learning for optimal execution. *Applied Mathematical Finance* 28(4):361–380, URL http://dx.doi.org/10.1080/1350486X.2022.2077783.

Pandey S, Karypis G (2019) A self-attentive model for knowledge tracing. *arXiv preprint arXiv:1907.06837* .

Pavlik Jr PI, Anderson JR (2005) Practice and forgetting effects on vocabulary memory: An activation-based model of the spacing effect. *Cognitive science* 29(4):559–586.

Peris-Ortiz M, Lindahl JMM (2015) *Sustainable learning in higher education* (Springer).

Piech C, Bassen J, Huang J, Ganguli S, Sahami M, Guibas LJ, Sohl-Dickstein J (2015) Deep knowledge tracing. *Advances in neural information processing systems* 28.

Quizlet(2024)Thesciencebehindspacedrepetitionlearning.Quizletblog,activerecallandspacedrepetitionexplained.

Radvansky GA, Doolen AC, Pettijohn KA, Ritchey M (2022) A new look at memory retention and forgetting. *Journal of Experimental Psychology: Learning, Memory, and Cognition* 48(11):1698.

Raghuveer V, Tripathy B, Singh T, Khanna S (2014) Reinforcement learning approach towards effective content recommendation in mooc environments. *2014 IEEE international conference on MOOC, innovation and technology in education (MITE)*, 285–289 (IEEE).

Reich J, Ruiperez-Valiente JA (2019) The mooc pivot.´ *Science* 363(6423):130–131.

Rezaee AA, Seyri H (2022) Curbing boredom in online teaching: Effects of an autonomy-oriented intervention. *Frontiers in Psychology* 13:1060422.

Sense F, Behrens F, Meijer RR, van Rijn H (2016) An individual's rate of forgetting is stable over time but differs across materials. *Topics in cognitive science* 8(1):305–321.

Settles B, Meeder B (2016) A trainable spaced repetition model for language learning. *Proceedings of the 54th annual meeting of the association for computational linguistics (volume 1: long papers)*, 1848–1858.

Su J, Ye J, Nie L, Cao Y, Chen Y (2023) Optimizing spaced repetition schedule by capturing the dynamics of memory. *IEEE Transactions on Knowledge and Data Engineering* 35(10):10085–10097.

Sun FY, Hoffmann J, Verma V, Tang J (2019) Infograph: Unsupervised and semi-supervised graph-level representation learning via mutual information maximization. *arXiv preprint arXiv:1908.01000* .

Todri V, Ghose A, Singh PV (2020) Trade-offs in online advertising: Advertising effectiveness and annoyance dynamics across the purchase funnel. *Information Systems Research* 31(1):102–125.

Tong S, Liu Q, Huang W, Hunag Z, Chen E, Liu C, Ma H, Wang S (2020) Structure-based knowledge tracing: An influence propagation view. *2020 IEEE international conference on data mining (ICDM)*, 541–550 (IEEE).

Tyler-Smith K (2006) Early attrition among first time elearners: A review of factors that contribute to drop-out, withdrawal and non-completion rates of adult learners undertaking elearning programmes. *Journal of Online Learning and Teaching* 2(2):73–85.

Vaswani A, Shazeer N, Parmar N, Uszkoreit J, Jones L, Gomez AN, Kaiser L, Polosukhin I (2017) Attention is all you need. *Advances in neural information processing systems* 30.

Velickovic P, Fedus W, Hamilton WL, Lio P, Bengio Y, Hjelm RD (2019) Deep graph infomax.` *ICLR (Poster)* 2(3):4. Von Glasersfeld E (2012) A constructivist approach to teaching. *Constructivism in education*, 3–15 (Routledge).

Wang Q, Mao Z, Wang B, Guo L (2017) Knowledge graph embedding: A survey of approaches and applications. *IEEE transactions on knowledge and data engineering* 29(12):2724–2743.

Wang W, Li B, Luo X, Wang X (2023) Deep reinforcement learning for sequential targeting. *Management Science* 69(9):5439–5460.

Wang W, Zhou M, Li B, Zhuang H (2025) Predicting instructor performance in online education: An interpretable hierarchical transformer with contextual attention. *Information Systems Research* .

Wang Y, Cai W, Chen M, Shen J (2020) Poem: a personalized online education scheme based on reinforcement learning. *2020 IEEE International Conference on Teaching, Assessment, and Learning for Engineering*, 474–481 (IEEE).

Xiao Q, Wang J (2024) Drl-srs: A deep reinforcement learning approach for optimizing spaced repetition scheduling. *Applied Sciences* 14(13):5591.

Xiong X, Zhao S, Van Inwegen EG, Beck JE (2016) Going deeper with deep knowledge tracing. *International Educational Data Mining Society* .

Xu D, Jaggars SS (2014) Does online learning affect student academic performance? evidence from a large community college system. *Economics of Education Review* 33:68–79.

Yang Y, Shen J, Qu Y, Liu Y, Wang K, Zhu Y, Zhang W, Yu Y (2020) Gikt: a graph-based interaction model for knowledge tracing. *Joint European conference on machine learning and knowledge discovery in databases*, 299–315 (Springer).

Ye J, Su J, Cao Y (2022) A stochastic shortest path algorithm for optimizing spaced repetition scheduling. *Proceedings of the 28th ACM SIGKDD conference on knowledge discovery and data mining*, 4381–4390.

## Appendix A: Notation Table

**Table 3 Variable Notation Summary**

| Symbol | Description |
|---|---|
| **Knowledge Graph** | |
| $K$ | The predefined set of all knowledge items, $\{k_1,\ldots, k_n\}$. |
| $t$ | The time step of a learning interaction. |
| $V$ | The set of nodes in the graph, corresponding to knowledge items. |
| $E$ | The set of directed edges representing relationships between items. |
| $G = (V, E)$ | The directed knowledge graph representing the learning space. |
| $e^{cd}_{i,j}$ | A directed curriculum-driven edge from knowledge item $i$ to $j$. |
| $e^{ed}_{i,j}$ | A directed empirically-driven edge from knowledge item $i$ to $j$. |
| $g_t$ | The subgraph of $G$ induced by items the student has learnt before time $t$. |
| **Reinforcement Learning** | |
| $s_t$ | The learner's state representation at time $t$. |
| $a_t$ | The action taken at time $t$, corresponding to the recommended knowledge item. |
| $r_t$ | The immediate reward received after the interaction at time $t$. |
| $A_t$ | Feasible action space at time $t$ |
| $\pi(s_t)$ | The policy function that maps a state to an action. |
| $Q(s_t, a_t)$ | The Q-value function, estimating the expected cumulative reward for an action in a state. |
| $\mathbf{u}_{a_t}$ | The embedding vector of node $a_t$ (i.e., knowledge item $a_t = k_i$). |
| $\mathbf{u}_{g_t}$ | The embedding vector of subgraph $g_t$, representing the student's accumulated knowledge state before time $t$. |
| $\mathbf{l}_t$ | A vector representing the learning interaction at time $t$, composed of $\{\mathbf{u}_{g_t}, \mathbf{u}_{a_t}, \mathbf{lp}_{a_t}\}$. |
| $\mathbf{lp}_{a_t}$ | A feature vector of the learning profile for item $a_t$ (e.g., study count, time since last seen). |
| $L_t$ | The learning trajectory, a sequence of interactions $\{\mathbf{l}_1,\ldots,\mathbf{l}_t\}$ up to time $t$. |
| $D$ | The replay buffer storing transition tuples $(s_t, a_t, r_t, s_{t+1})$. |
| $r^{\mathrm{acq}}_t$ | The reward component from knowledge acquisition at time $t$. |
| $r^{\mathrm{ret}}_t$ | The reward component from knowledge retention (memory strengthening) at time $t$. |
| $p^{\mathrm{rec}}$ | The predicted probability that the student can recall item $a_t$ before the interaction. |
| $t$ | |
| $\beta$ | A hyperparameter to scale the retention reward $r_t^{\text{ret}}$. |
| $h_{\text{pre}(t)}$, $h_{\text{post}(t)}$ | The half-life strength of an item's memory trace before and after an interaction. |
| $\Delta t$ | The time interval since an item was last studied/reviewed. |
| $L_{\text{critic}}(\theta)$, $L_{\text{actor}}(\phi)$ | The loss function for the critic network and the actor network. |
| **Learner Behavior Simulator** | |
| $p_t^{\text{eng}}$ | The predicted probability that the student will continue learning after interaction $t$. |
| $\text{eng}_t$ | The ground truth label for learner engagement (1 if continues, 0 otherwise). |
| $\text{rec}_t$ | The ground truth label for recall of item $a_t$ (1 if recalled, 0 otherwise). |
| $\boldsymbol{M}_t$ | The attention memory tensor from the LearnSim structure at time $t$. |
| $L_{\text{LearnSim}}$ | The loss function of model LearnSim |
| $\text{Loss}_{\text{eng}}$, $\text{Loss}_{\text{rec}}$, $\text{Loss}_{\text{ret}}$ | The loss component for engagement prediction, recall probability prediction, half-life (retention) prediction |
| $\lambda_1$, $\lambda_2$ | Hyperparameters to weight the loss components in $L_{\text{LearnSim}}$. |

## Appendix B: Empirical Context - User Learning Jounery and APP Interface

The user's learning journey on MaiMemo follows a structured and interactive process. Each learning course consists of a specific number of required daily sessions. Each session includes multiple interactions, with each interaction focusing on a single vocabulary item. This vocabulary item may be new or previously learned by the user.

As illustrated in Figure 12, at the start of each interaction, the app deliberately withholds the word's definition and instead may present one or more example sentences to stimulate contextual recall. The learner then evaluates their familiarity with the word by choosing one of three options: "known," "uncertain," or "forgotten." Based on this self-assessment, the app subsequently reveals comprehensive learning materials, including the word's definition, usage examples, and related content to reinforce understanding. Such interactions occur repeatedly, often multiple times (e.g., ) in a single day's learning session. The system schedules each word's future learning sequence based on the learner's current session and the performance history of previous sessions. This process is guided by an internal learning path recommendation algorithm.

It is worth noting that, as with many online learning platforms, users on MaiMemo may disengage from the learning process at any point, resulting in incomplete sessions or interrupted interactions. A learner may choose to pause mid-session and return later the same day to resume learning, initiate a new session on a subsequent day, or, in some cases, abandon the platform altogether. MaiMemo defines user dropout as a case in which the learner does not return to the platform for at least 14 consecutive days, a behavioral pattern commonly observed across digital education environments.

This learning journey exemplifies a sequential and iterative learning path, a structure widely adopted across online education platforms. Such a structure typically involves progressive exposure to instructional content, followed by learner evaluation and feedback, forming a continuous loop that facilitates both acquisition and long-term retention of knowledge. This paradigm is evident across various educational platforms: for instance, memorization-based applications like Quizlet emphasize repetitive recall of individual knowledge units, while comprehension-focused platforms such as Coursera and edX deliver structured course content accompanied by formative assessments (e.g., quizzes, exercises, or peer reviews). The learning pattern observed in MaiMemo mirrors these broader instructional frameworks, making it a representative case of sequential digital learning experiences.

## Appendix C: Model Training and Implementation

### C.1. Knowledge Graph Construction

In our empirical setting, our approach to constructing a knowledge graph for English language learning involves defining each vocabulary word as a node and connecting them with two types of edges: *curriculum-driven* and *empirically-driven*.

*Curriculum-Driven* These edges represent explicit semantic relationships derived from authoritative sources like the Collins and Oxford English dictionaries. An edge is drawn between two vocabulary nodes if they share one of the following relationships: 1. Word families – sets of words that share a common root along with their inflections or derivations (e.g., teach, teacher, teaching). Learning words through families enables learners to recognize recurring patterns and infer meanings more effectively, particularly when they acquire the root first. 2. Synonym (or Antonym) Sets - groups of words that have similar or identical meanings, or opposite meanings. For example, synonyms include

**Figure 12** Example User Interface from an Educational Application

8:59
impulse
美 [ɪmpʌls]
Knowledge Item
请回忆单词发音和释义
点击屏幕显示答案
User Response: known, uncertain, forgotten
认识
138 天后
模糊
今日 / 1 天后
忘记
今日 / 1 天后
复习
选词
统计
我的

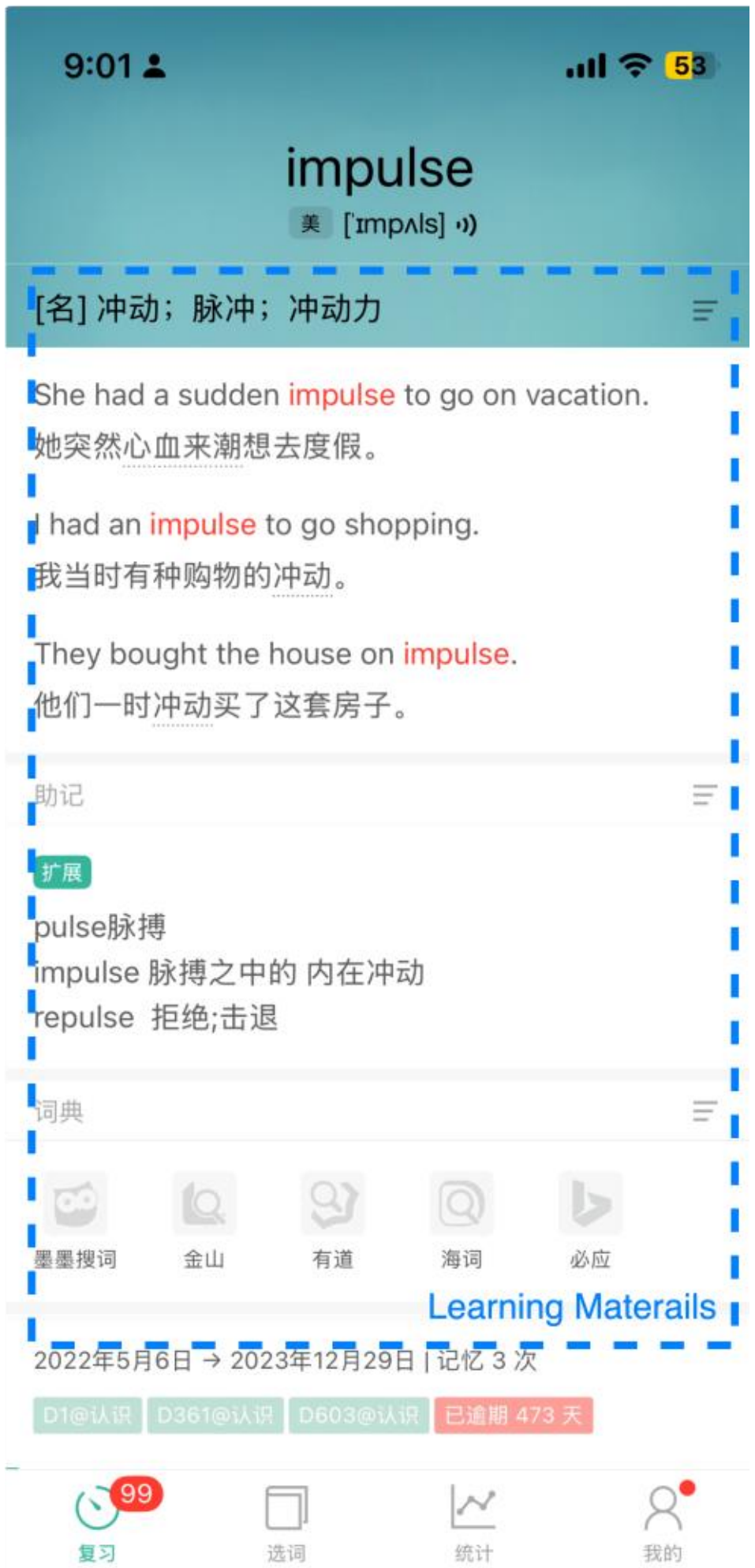

*Note.* The left panel illustrates a typical interaction screen where users are prompted to recall a concept based on contextual cues. The right panel displays the detailed explanation and feedback following the user's input.

"big," "large," and "sizable." This helps learners expand their expressive range while reinforcing subtle distinctions in meaning and usage.

Such connections highlight the semantic closeness and learning prerequisites of words, thereby making them easier for learners to link, internalize, and understand. Our knowledge graph framework supports directed relationships. For undirected relationships, such as synonym or antonym, we treat the edges as bidirectional. This means the knowledge can be prerequisites for either entity involved.

*Empirically-Driven* To discover implicit, complementary learning patterns, we create empirically-driven edges by applying association rule mining to our large dataset of student learning records. This helps us identify words that students tend to master together. We use the following standard metrics to define directed relationships ($i \rightarrow j$):

- *Support ($i$):* Measures how frequently is word $i$ mastered once it has been learned.

$$\text{Support}(i) = \frac{\text{\# of learning interactions where learners mastered } i}{\text{Total \# of learning interactions including } i}$$

- *Support ($i, j$):* Measures how frequently words $i$ and $j$ are mastered together by the same learner.

$$\text{Support}(i, j) = \frac{\text{\# of learning interactions where learners mastered both } i \text{ and } j}{\text{Total \# of learning interactions including } i \text{ and } j}$$

- *Confidence ($i \rightarrow j$):* Indicates the probability that a learner will master word $j$ given they have mastered word $i$. This establishes a *directed edge* from $i$ to $j$.

$$\text{Confidence}(i \rightarrow j) = \frac{\text{Support}(i, j)}{\text{Support}(i)}$$

- *Lift ($i \rightarrow j$):* Measures how much more likely a learner is to master word $j$ given they have mastered word $i$, compared to mastering $j$ by chance.

$$\text{Lift}(i \rightarrow j) = \frac{\text{Confidence}(i \rightarrow j)}{\text{Support}(j)}$$

To build the graph, we first filter for edges where the *Lift* is greater than 1, which signifies a positive, non-random dependency. Next, we set a minimum *Confidence* threshold, initially starting with the mean confidence value of all potential edges. We then incrementally raise this threshold to retain only the strongest connections, while ensuring the graph remains fully connected.

The final output is a robust knowledge graph that integrates both explicit semantic relationships and statistically significant learning dependencies to support English language learning.

## C.2. Knowledge Graph Embedding

**C.2.1. DGI for the Representation Learning of Knowledge Item** For this part, we employ a graph representation learning model to obtain node embeddings on the constructed KG. Specifically, we adopt the state-of-the-art unsupervised method Deep Graph Infomax (DGI)((Velickovic et al. 2019)). The core idea behind DGI is to teach the model to understand the authentic structure of our knowledge graph by learning to distinguish it from a corrupted version of itself. To create this "corrupted" version, the attributes of the nodes are randomly shuffled, but the original network of connections (i.e., the original adjacency structure) is preserved. The model is then trained to identify which nodes truly belong to the original graph structure.

Specifically, the DGI model learns an encoder $f_{DGI}$ that generates embeddings for nodes in the original graph, $\mathbf{H} = \{\mathbf{h}_i\}_{i=1}^{N}$, and for nodes in the corrupted graph, $\tilde{\mathbf{H}}$. It also computes a single summary vector, $\mathbf{s}$, which represents an aggregated view of the entire graph (e.g., $\mathbf{s} = \sigma(\frac{1}{N}\sum_{i=1}^{N}\mathbf{h}_i)$). The model is then trained to produce a high score for real pairs $(\mathbf{h}_i, \mathbf{s})$—indicating a good fit—and a low score for corrupted pairs $(\tilde{\mathbf{h}}_i, \mathbf{s})$.

$$\mathcal{L}_{DGI} = -\frac{1}{2N}\sum_{i=1}^{N}\left[\log\sigma(\mathbf{h}_i^{\top}\mathbf{W}\mathbf{s}) + \log(1-\sigma(\tilde{\mathbf{h}}_i^{\top}\mathbf{W}\mathbf{s}))\right] \tag{15}$$

By successfully telling these pairs apart, the model implicitly learns the underlying rules that govern how knowledge items are structured in the graph. The result is a set of final embeddings that are both distinct and contextually aware. We obtain the final embedding for each knowledge item $i$ from the trained encoder:

$$\mathbf{u}_{k_i} = f_{DGI}(i, G) \tag{16}$$

**C.2.2. DGI for the Representation Learning of Learning Path** To better capture the learner state, we identify the learner's learning path $g_t$ on the knowledge graph $G$ at each time $t$.

The learning path $g_t$ is a directed subgraph of $G$ that tracks the number of times a learner has engaged with each knowledge item $i$, along with their most recent learning performance up to time step $t$. In math, that is $g_t = (V_t, E_t)$, where $V_t = \{(i, c_{a_t}, p^{rec}_t) \mid i \in G\}$ is the set of knowledge items with their historical learning counts $c_{a_t}$ and learning performance $p^{rec}_t$, and $E_t \in G$ represents the directed edges between knowledge items within the learning path.

We then use InfoGraph (Sun et al. 2019), a graph-level embedding approach extended from DGI. The objective of InfoGraph is to generate a single vector representation for each learning path, such that paths with similar structures or concepts are mapped to nearby points in the embedding space. Following the same principle as DGI, it is trained to recognize that the nodes within a learning path are authentic components, while nodes sampled from other learning paths are not.

To achieve this, the InfoGraph encoder, $f_{IG}$, first computes embeddings for all nodes within a given learning path, $\mathbf{H}_{g_t}$. These are then aggregated via a readout function to produce a single subgraph embedding, $\mathbf{s}_{g_t}$ (e.g., $\mathbf{s}_{g_t} = \sigma(\frac{1}{|V_t|}\sum_{i \in V_t}\mathbf{h}_i)$). The model's contrastive task is then defined at the subgraph level. For a given learning path $g_t$, a positive pair consists of one of its own node embeddings ($\mathbf{h}_i$) and its overall subgraph embedding ($\mathbf{s}_{g_t}$). A negative pair is formed by contrasting that same subgraph embedding with a node embedding ($\tilde{\mathbf{h}}_j$) sampled from a different learning path. The loss function encourages the model to assign high similarity scores to positive pairs and low scores to negative pairs:

$$\mathcal{L}_{IG} = -\frac{1}{|V_t|} \sum_{i \in V_t} \left[ \log \sigma(\mathbf{h}_i^{\top} \mathbf{W} \mathbf{s}_{g_t}) + \log(1 - \sigma(\tilde{\mathbf{h}}_j^{\top} \mathbf{W} \mathbf{s}_{g_t})) \right] \tag{17}$$

By training the model to distinguish its internal nodes from external ones, the InfoGraph encoder learns to produce a final subgraph embedding, $\mathbf{u}_{g_t}$, that encapsulates the essential properties of the entire learning path.

$$\mathbf{u}_{g_t} = f_{IG}(g_t, G) \tag{18}$$

Later, the subgraph embedding $\mathbf{u}_{g_t}$ will be processed through RNN-based layers in our reinforcement learning model and prediction model to capture the sequential patterns.

### C.3. LearnSim Training

The dataset used in this study is composed of over 23 million study records from 33,700 unique users. To facilitate model training and evaluation, the data was partitioned at the user level, ensuring that all records from a single user belong exclusively to one set. We allocated 40% of users for the training set, 10% for the validation set, and 10% for the final test set. The rest 40% was employed to train a separate LearnSim for evaluating AI-Tutor.

**C.3.1. Hyperparameter Tuning** Hyperparameter optimization was conducted using a two-stage, coarse-to-fine strategy for computational efficiency. Initially, promising hyperparameter ranges were identified through manual tuning. Subsequently, a grid search was performed on these ranges using the validation set. The AUC-ROC was the primary metric for selecting the optimal hyperparameter configuration. We also track other metrics, including Log Loss, accuracy, F1-Score, average precision, and balanced accuracy. The complete search space and final optimal values are detailed in Table 4.

The model was trained using an NVIDIA A100 GPU, with a 32-core processor and 128 GB of RAM. The Python version used is 3.11.9 and PyTorch version is 2.3.0.

**Table 4** Hyperparameter Configuration for the LearnSim Prediction Model

| Hyperparameter | Definition | Search Space | Optimal Value |
|---|---|---|---|
| **Knowledge Graph-related** | | | |
| KG Compress Dims | The dimension of the KG embedding after the compression layer. | 8 | 8 |
| Embedding Size | The final input embedding size for the Transformer model. | {16, 32, 64} | 32 |
| **Transformer-related** | | | |
| Num Layers | Number of sequential Transformer encoder layers. | {1, 2, 4} | 2 |
| Num Heads | Number of heads in the multi-head attention mechanism. | {2, 4, 8} | 4 |
| Dropout Rate | Dropout probability for regularization. | {0.0, 0.1, 0.2} | 0.1 |
| $\lambda_1, \lambda_2$ | Hyperparameters to weight multi-task objectives in loss function | {0.3,0.6,0.8,1,1.2} | 1, 0.3 |
| **Training Parameters** | | | |
| Learning Rate | Learning Rate for the optimizer. | $\{10^{-5}, 10^{-4}, 10^{-3}\}$ | $10^{-4}$ |
| Batch Size | Number of sequences per training batch. | {8, 16, 32} | 16 |
| Optimizer | Optimization algorithm. | Adam | Adam |
| Num Epochs | The maximum number of training epochs. | 1000 | 1000 |
| Patience | Epochs to wait for improvement before early stopping. | {10, 20, 30} | 30 |
| Min Delta | Minimum required change to be considered an improvement. | 0.001 | 0.001 |

**C.3.2. Benchmarks Training** To evaluate our proposed model, we compared its performance against several established baselines and an ablation variant of our own model. For all benchmarks, we used consistent training, validation, and test splits to ensure a fair comparison, and we conducted basic hyperparameter tuning.

- *HLR (Half-Life Regression)*: This model, proposed by (Settles and Meeder 2016), is a widely-used memory model that forms the basis of Duolingo's spaced repetition system. The original official implementation is available at https://github.com/duolingo/halflife-regression. To accommodate the high-dimensional input features, we refer to the neural network variant code, which is available at https://github.com/ open-spaced-repetition/srs-benchmark/blob/main/models/hlr.py. The model is trained using the Adam optimizer with a learning rate of $1 \times 10^{-2}$, a batch size of 128, and a maximum of 1000 epochs. We apply early stopping based on the validation set performance to prevent overfitting.

- *DHP-HLR*: An extension of HLR that incorporates a Markov property to model learning dynamics (Ye et al. 2022).

We adapt the official source code from https://github.com/maimemo/SSP-MMC/blob/main/model/ dhp.py for our experiments. We follow the hyperparameter settings from the original paper, training the model with the Adam optimizer, a learning rate of $1 \times 10^{-3}$, and a batch size of 128, and a maximum of 1000 epochs. We apply early stopping based on the validation set performance to prevent overfitting.

- *GRU-HLR*: This model replaces the static state transitions in HLR with a Gated Recurrent Unit (GRU) to learn memory state updates automatically (Su et al. 2023). Our implementation is based on the official implementation code, available at https://github.com/maimemo/SSP-MMC-Plus/blob/main/model/RNN_HLR.py. The model is trained using the Adam optimizer with a learning rate of $1 \times 10^{-3}$, a batch size of 128, and a maximum of 1000 epochs. We apply early stopping based on the validation set performance to prevent overfitting.

- *LearnSim$_{v1}$*: This is an ablation study variant of our full model. It is designed to isolate the contribution of our knowledge graph component by replacing the KG-based embeddings with 8-dimensional Word2Vec embeddings (Mikolov et al. 2013). This variant uses the exact same architecture and hyperparameter tuning protocol as our full LearnSim model, with the sole exception of the input word representations.

### C.4. AI-Tutor Training

**C.4.1. Soft Actor-Critic Structure**In our practical implementation, to enhance stability and mitigate overestimation, we draw inspiration from Soft Actor-Critic (Haarnoja et al. 2018).

Specifically, for critic, we utilize two critic networks ($Q_1(s_t, a_t \mid \theta_1)$ and $Q_2(s_t, a_t \mid \theta_2)$) and their corresponding target networks ($Q_{1,\text{target}}$ and $Q_{2,\text{target}}$). The target value $y_t$ for updating both critics takes the minimum of the two target Q-network estimations. Furthermore, to encourage broader exploration, our target value incorporates an entropy-based
term:

$$y_t = r_t + p_t \mathrm{E}^{\text{eng}}_{s_{t+1}, a_{t+1} \sim \pi} \min_{j=1,2} \{Q_j(s_{t+1}, a_{t+1})\} + \omega \mathrm{H} \tag{19}$$

Here, $a_{t+1} \sim \pi(\cdot \mid s_{t+1})$ (i.e., sampled from the current actor policy for the next state $s_{t+1}$). $\mathrm{H} = -\log \pi(a_{t+1} \mid s_{t+1})$ is the entropy term associated with the next action. By incorporating entropy, the algorithm favors policies that remain stochastic, compelling the agent to attempt diverse actions to encourage exploration. $\omega > 0$ is a self-learnable coefficient
balancing the reward and entropy (Haarnoja et al. 2018). Each online critic $Q_j(s_t, a_t \mid \theta_{cj})$ is then updated using the loss $\mathrm{L}_{\text{critic}}(\theta)$ in Equation (8) with this target $y_t$.

To align with the critic's learning objective that includes an entropy component, the actor aims to maximize both the expected Q-value of its actions and the entropy of its action distribution. The objective for updating the policy $\pi(\cdot \mid s_t; \phi)$ is thus:

$$\mathrm{L}_{\text{actor}}(\phi) = \mathrm{E}_{s_t \sim D, a_t \sim \pi_\phi(\cdot \mid s_t)} \min_{j=1,2} \{Q_j(s_t, a_t\} + \omega \mathrm{H}' \tag{20}$$

where $a_t$ is an action sampled from the policy $\pi(\cdot \mid s_t; \phi)$ for state $s_t$ (from the replay buffer), $\min_{j=1,2}\{Q_j(s_t, a_t \mid \phi_j)\}$ utilizes the minimum Q-value estimate from our two online critic networks, and $\mathrm{H}' = -\log \pi_\phi(a_t \mid s_t)$ is the entropy term for the chosen action. The policy parameters $\phi$ are typically updated by ascending the gradient of this objective.

#### C.4.2. Other Practical Implementation Setups

*Environment Initialization* The initial states of the simulated environment (learners) are randomly sampled from real-world study records on the *MaiMemo* platform. In practice, this means sampling the first 20 study records as the initial state for each environment.

*KG-based Action Masking* To optimize our KG-based action masking mechanism and improve model training efficiency, we refine how the action space is defined. In practice, considering a learner's entire history is computationally intensive and can introduce noise from long-mastered topics. Therefore, we define a dynamic *learning path* $L_t$, at each time step $t$. This path is composed of the $n$ most recent knowledge items the learner has mastered, providing a more accurate reflection of their current knowledge frontier.

Using this learning path, we construct a localized feasible action space, $A_t$, consisting of the $m$ nearest neighbors to the nodes in $L_t$ on the Knowledge Graph. The parameters $n$ and $m$ are hyperparameters that balance exploration and efficiency: A large $m$ broadens the search for the next topic but diminishes the guiding precision of the KG mask. Conversely, a small $m$ creates an overly restrictive candidate set, risking the omission of optimal learning actions.

To find the ideal configuration, we select the optimal values for $n$ and $m$ through systematic hyperparameter tuning via grid search.

*Hardware and Software Environments* All AI-Tutor models, including benchmarks, are trained and tested on an Apple M2 with 8 cores and 24GB of RAM. The Python version used is 3.11.9. The key Machine Learning and Reinforcement Learning packages used are : PyTorch: 2.3.0, Tianshou: 1.0.0, Gymnasium: 0.28.1.

**C.4.3. Hyperparameter Tuning** For hyperparameter tuning, we begin with manual tuning. Our initial goal is to identify a rough "trusted area," after which we employ Grid Search for fine-tuning. Detailed hyperparameter tuning information can be found in Table 5.

**Table 5 AI-Tutor's Hyperparameter Configuration**

| Hyperparameter | Definition | Search Space | Optimal Value |
|---|---|---|---|
| **Task-Specific** | | | |
| $\beta$ | A coefficient to scale the retentionawareness reward. | {0.05, 0.1, 0.2, 0.5, 0.7, 1} | 0.1 |
| $n$ | Number of the most recent items considered in the learning path. | {1, 5, 10, 20} | 10 |
| $m$ | Number of nearest neighbors for the $n$ most recent items. | {10, 30, 50, 100} | 30 |
| **SAC-related** | | | |
| Replay Buffer Size | Number of transitions stored in replay memory. | $\{10^5, 10^6, 10^7\}$ | $10^6$ |
| Batch Size | Numberofsamplespergradientupdate. | {8, 16, 32} | 16 |
| Learning Rate | Learning rate for actor, critic, and $\omega$. | $\{10^{-5}, 10^{-4}, 10^{-3}\}$ | $10^{-4}$ |
| Polyak Rate ($\tau$) | Target network soft update rate. | {0.005, 0.01} | 0.005 |
| Optimizer | Optimizer for all networks. | Adam | Adam |
| Actor/Critic Architecture | Hidden size and layers for the actor/critic MLPs. | Size: {128, 256}, Layers: { 2, 4} | {256, 2} |
| RNN Architecture | LSTM hidden size and layers for state encoding. | Size: {64, 128}, Layers: {1} | {64, 1} |
| Entropy Coeff. ($\omega$) | Balances the reward and policy entropy in the SAC objective. | Automatically tuned (start from 0.2) | Learnable |

**C.4.4. Benchmarks Training** We evaluate our proposed AI-Tutor against a diverse set of baselines, including heuristic methods, a rule-based system, a state-of-the-art deep reinforcement learning model, and several ablation variants of our own model.

- *MyopicGreedy*:Apurelyexploitativebaselinethatalwaysselectstheitemwiththelowestcurrentrecallprobability. This method is parameter-free and does not require training.
- *FSRS$_{v1}$*: A widely adopted rule-based scheduling system (FSRS 2024) that recommends items for review when their estimated retention falls below a predefined threshold (0.5 in our experiments). Following standard practice, we use a Half-Life Regression (HLR) model (Settles and Meeder 2016) to estimate memory strength. As a rule-based method, it does not require training beyond the underlying HLR model. Our implementation follows the logic from the official repository: https://github.com/open-spaced-repetition/fsrs4anki.
- *FSRS$_{v2}$*: This variant maintains the same scheduling logic as FSRS$_{v1}$ but replaces the HLR memory model with our pre-trained LearnSim model to estimate memory strength. This allows us to leverage LearnSim's superior predictive performance in a rule-based context. This method requires no separate training.
- *DRL-SRS*: This baseline is a deep reinforcement learning framework based on Deep Q-Networks (Xiao and Wang 2024). As official source code was not available, we re-implemented the model following the architectural descriptions and procedures outlined in the original paper. The model was trained using the Adam optimizer with a learning rate of $1 \times 10^{-4}$ and a replay buffer of size $10^6$.
- *AI-Tutor$_{v1}$*: To assess the impact of engagement awareness, this ablation model replaces the dynamic churn probability $p^{eng}_t$ in the Q-function (Equation 7) with a fixed discount factor of $\gamma = 0.99$.
- *AI-Tutor$_{v2}$*: This version evaluates the contribution of our structural guidance mechanism by disabling KG-based action masking.
- *AI-Tutor$_{v3}$*: To isolate the effect of the retention-focused reward, this model removes the knowledge retention component ($r_t^{ret}$) from the reward function. The agent is thus trained solely on the acquisition reward, $r_t = r_t^{acq}$.

For all ablation models (AI-Tutor$_{v1}$, AI-Tutor$_{v2}$, and AI-Tutor$_{v3}$), all other aspects, including architecture and the hyperparameter tuning protocol, are identical to the full AI-Tutor model to ensure a controlled comparison.

**C.4.5. ComplexityAnalysis** WeadoptedtheSACstructureinpracticalimplementation.Belowisabriefcomplexity analysis from both theoretical and practical perspectives.

*Theoretical Complexity Analysis* Theoretically,thetimecomplexityofasingleSoftActor-Critic(SAC)update step is $O(N \cdot (C_A + C_C))$. In this equation, $N$ represents the batch size, which is the number of data samples processed in one update. The terms $C_A$ and $C_C$ correspond to the computational cost of a single forward pass through the actor (policy) network and one critic (Q-value) network, respectively. The complexity of a neural network isn't a simple number. For a standard feed-forward network (MLP), the complexity of a forward pass is the sum of the complexities of its matrix multiplications. For a single layer with $d_{in}$ inputs and $d_{out}$ outputs, the complexity is $O(d_{in} * d_{out})$. So, $C_A(C_C)$ is actually $\sum O(d_i * d_{i+1})$ for all layers in the actor(critic). Essentially, the training time for SAC scales linearly with both the amount of data you use in each step and the architectural complexity (i.e., size and depth) of your neural networks.

*Practical Complexity Analysis* Practically, our implementation leverages multi-processing to accelerate training by enabling the agent to interact with *8 parallel environments* simultaneously. All experiments were conducted on a machine with an Apple M2 (8-core) CPU and 24GB of RAM, using Python 3.11.9, PyTorch 2.3.0, Tianshou 1.0.0, and Gymnasium 0.28.1. This setup demonstrated high efficiency and consistent performance, as detailed below:

- *Throughput of Training:* 190 interactions per second, with an average interaction time of 5.26 ms.
- *Throughput of Inference:* 310 interactions per second, with an average interaction time of 3.23 ms.

Overall, this efficient performance confirms the system's suitability for conducting large-scale reinforcement learning experiments within reasonable timeframes and its feasibility for real-time responses in real-world applications.